\documentclass[twoside,11pt]{article}
\usepackage[utf8]{inputenc}
\usepackage{jair, theapa, rawfonts}

\usepackage{amsmath}
\usepackage{amssymb}
\usepackage{amsthm}
\usepackage{mathtools}
\usepackage{bbm}
\usepackage{multirow}
\usepackage{graphicx}
\graphicspath{{Figures/}}
\usepackage[linesnumbered,algoruled,vlined]{algorithm2e}
\usepackage[caption=false]{subfig}
\usepackage{tikz}
\usepackage{tikz-qtree}
\usetikzlibrary{positioning,angles,quotes,shapes.misc}

\newtheorem{theorem}{Theorem}
\newtheorem{lemma}{Lemma}

\DeclareMathOperator*{\argmax}{argmax}

\SetKwFor{Loop}{loop}{}{end}

\title{Vulcan: A Monte Carlo Algorithm for Large Chance Constrained MDPs with Risk Bounding Functions}

\author{\name Benjamin J. Ayton \email aytonb@mit.edu \\
       \name Brian C. Williams \email williams@mit.edu \\
       \addr Computer Science and Artificial Intelligence Laboratory, Massachusetts Institute of Technology\\
       32 Vassar St, Cambridge MA, 02139}

\begin{document}

\maketitle

\begin{abstract}
Chance Constrained Markov Decision Processes maximize reward subject to a bounded probability of failure, and have been frequently applied for planning with potentially dangerous outcomes or unknown environments. Solution algorithms have required strong heuristics or have been limited to relatively small problems with up to millions of states, because the optimal action to take from a given state depends on the probability of failure in the rest of the policy, leading to a coupled problem that is difficult to solve. In this paper we examine a generalization of a CCMDP that trades off probability of failure against reward through a functional relationship. We derive a constraint that can be applied to each state history in a policy individually, and which guarantees that the chance constraint will be satisfied. The approach decouples states in the CCMDP, so that large problems can be solved efficiently. We then introduce Vulcan, which uses our constraint in order to apply Monte Carlo Tree Search to CCMDPs. Vulcan can be applied to problems where it is unfeasible to generate the entire state space, and policies must be returned in an anytime manner. We show that Vulcan and its variants run tens to hundreds of times faster than linear programming methods, and over ten times faster than heuristic based methods, all without the need for a heuristic, and returning solutions with a mean suboptimality on the order of a few percent. Finally, we use Vulcan to solve for a chance constrained policy in a CCMDP with over $10^{13}$ states in 3 minutes.
\end{abstract}

\section{Introduction}

Markov Decision Processes (MDPs) are a powerful framework for artificial intelligence systems that must perform optimally in the face of uncertainty. MDPs have been broadly applied in various domains, ranging from inventory control to communication to motion planning, and algorithms for finding optimal policies are mature~\cite{puterman2014}. However, in many circumstances reward is not the only consideration for a policy. For instance, when an autonomous agent operates in an uncertain environment, actions have a probability of collision with obstacles which jeopardizes the future of the mission. In such circumstances, it is undesirable to simply maximize a measure of reward because high reward policies can be tied to dangerous actions, leading to unacceptably large chances of failure.

Chance Constrained MDPs (CCMDPs) place a constraint on the allowed probability of failure in the policy, which we refer to as either the chance constraint or the risk bound~\cite{rossman1977}. The addition of the risk bound complicates solving the CCMDP since it is no longer optimal to select the highest cumulative reward action from each state, and known solution techniques do not scale well to very large problems. While Monte Carlo Tree Search (MCTS) based planning has shown remarkable recent successes in large MDPs, it has not been applied to CCMDPs because a chance constraint couples all actions and outcomes in a policy, and it cannot trivially be reasoned over in separate branches when the rest of the policy is unknown.

In addition, we find a traditional CCMDP that asserts a constant risk bound to be insufficiently expressive, as in reality the acceptable probability of failure is often contingent upon the reward that would be achieved. Intuitively, mission designers will accept a greater probability of failure if a riskier mission is likely to yield much more reward. But marginal risk tolerance also tends to shrink with increasing reward. This means a mission with a much larger probability of failure but a small increase in reward over an already high reward mission is unlikely to be preferred. Altogether this implies a risk bound that is a concave nondecreasing function of the mission reward, a generalization of static risk bounds. 

In this paper we introduce Vulcan, an MCTS based algorithm for large CCMDPS with a concave nondecreasing risk bounding function. We derive a sufficient condition that can be applied during Monte Carlo Tree Search so that any policy returned by the algorithm is guaranteed to satisfy a bound on probability of failure, computed as a function of expected reward of the policy. This allows the algorithm to be run in an anytime manner without the need to explore all states in the policy, at the cost of converging to a slightly suboptimal policy.  

Experiments with Vulcan on smaller problems where the optimal policy can be found suggest that the mean suboptimality is on the order of a few percent. Vulcan is observed to run between 50 and 600 times faster than methods that explicitly explore the state space, and for large problems it is observed to run over 10 times faster than heuristic forward search methods. Finally, we demonstrate the use of Vulcan to find a chance constrained policy in a CCMDP with approximately $3.5 \times 10^{13}$ states in 3 minutes.



\section{Motivating Scenario}

To motivate the development of Vulcan, consider an autonomous vehicle exploring an unknown environment in search of high reward samples, for example, an underwater vehicle exploring the oceans of Europa. As each sample is taken, it updates its model of the environment around it. The position estimate of the vehicle is a probability distribution through space, and when the vehicle's samples take it close to obstacles, there is a risk of collision that could damage the vehicle and end its mission. 

A low constant risk bound means that the vehicle will stay far away from obstacles, even if the most interesting samples are near them. A high constant risk bound is similarly undesirable, as the optimal policy will move the vehicle close to obstacles even if the samples are worth only slightly more reward. Instead, desirable behavior would include an interplay between risk and reward, in the sense that additional risk should be taken only if the additional reward is deemed worthwhile. A natural expression for this balance is a function that specifies the maximum allowable probability of failure for every expected reward. Vulcan allows any concave nondecreasing risk bounding function to be specified, and finds a satisfactory policy accordingly.     

In this case, the environment model that predicts the outcomes of future actions is a function of the locations and outcomes of \emph{all} previous samples. The probability of failure depends on the locations the vehicle visits, but also the order in which they are visited because the uncertainty in vehicle state typically grows with time. Since the optimal action to take depends on the possible outcomes of actions as well as the total reward and probability of failure incurred up to a state, the rewards and transition probabilities from a state (and therefore the optimal policy from that state) will generally depend on the entire history of states and actions preceding it. For even moderately sized environments, the set of states becomes very large. Vulcan is able to handle these large state spaces by sampling from them in an intelligent manner, which guides the search towards the final policy without evaluating all possible states. 

This exploration scenario has one more feature that was important for the development of Vulcan, which is the fact that rewards and risks are expensive to compute. Propagating a sample outcome through an environment model can be computationally difficult, especially for larger models, and reasoning over probability distributions to compute collisions can also be time consuming. The expense of evaluating states further justifies an MCTS approach that does not generate every state in the CCMDP. Vulcan also accounts for expensive states by storing results in memory where possible to avoid recomputation.

\section{Problem Statement}


We consider the problem of finding the optimal policy in a finite horizon CCMDP subject to a risk bounding function. Formally, a finite horizon CCMDP is a tuple $\langle \mathcal{S}, \mathcal{C}, \mathcal{A}, T, R, \gamma, s_{0}, n, \Delta \rangle$, where:
\begin{itemize}
    \item $\mathcal{S}$ is a set of states.
    \item $\mathcal{C} \subseteq \mathcal{S}$ is a set of \emph{safe states}, which satisfy mission constraints such as staying outside of obstacles, while states in $\mathcal{S} \setminus \mathcal{C}$ are considered \emph{failure states}.
    \item $\mathcal{A}$ is a set of actions.
    \item $T: \mathcal{S} \times \mathcal{A} \times \mathcal{S} \rightarrow [0,1]$ is a stochastic state transition function indicating the probability of transitioning from one state to another when taking an action.
    \item $R: \mathcal{S} \times \mathcal{A} \times \mathcal{S} \rightarrow \mathbb{R}$ is a reward function giving the numeric reward from moving between states according to an action.
    \item $\gamma \in \left[0,1\right]$ is a discount factor that prioritizes immediate reward, so that a reward received $t$ actions in the future is worth $\gamma^{t}$ times its original value.
    \item $s_{0} \in \mathcal{C}$ is an initial state, representing the state of the world before any actions are taken.
    \item $n$ is the planning horizon, or number of actions to perform in the CCMDP.
    \item $\Delta: \mathbb{R} \rightarrow \left[0,1\right]$ is a concave nondecreasing risk bounding function which gives the maximum acceptable probability of entering a failure state as a function of reward.
\end{itemize}
Missions where the initial state is chosen from a discrete set (for example, deploying a vehicle in different locations) or is distributed according to a probability distribution may be modeled with a fixed starting point using an initial action from a dummy state to decide the initial state.

We denote a state at time step $t$ as $s_{t}$ and a state history from time step $t$ to $t'$ as $h_{t:t'} = (s_{t}, a_{t}, s_{t+1}, a_{t+1}, \dots, s_{t'})$ (a sequence of states and actions between those states). The set of all possible state histories is denoted as $\mathcal{H}$, and the set of all possible state histories that include only safe states, which we call safe state histories, is denoted as $\mathcal{H}^{\mathcal{C}}$.  We seek a policy $\pi: \mathcal{H} \rightarrow \mathcal{A}$ to be followed such that the probability of entering an failure state is bounded according the risk bounding function $\Delta$.

While the optimal policy may not be deterministic in general, Vulcan will produce an approximately optimal deterministic policy. It has been argued that randomized policies are difficult to reliably execute, particularly when multiple agents are involved \cite{dolgov2005,paruchuri2004}. In addition, from a mission planning and tracking perspective, deterministic policies are easier to interpret and, if necessary, repeat. Regardless, even if a randomized policy \emph{is} permitted, we will show that Vulcan still provides order of magnitude speedups over solution methods which explore the entire state space. 

In the types of problems we will consider, a state depends strongly on its history, in the sense that a state will only be reachable by one or a small number of histories, like in our exploration scenario. Despite the fact that Vulcan uses a tree structure and considers each state history distinctly, we will show that it still performs well against techniques that do not repeat states when a state is reachable by several histories. However, in the extreme of problems with a small number of states with many loops and very long planning horizons, we would not expect the techniques introduced in this paper to perform well.

In this work, we consider failure to satisfy the constraints as disastrous in nature, such as damaging and losing the exploration vehicle or corrupting all data.\footnote{Strictly speaking, this is not a necessary assumption, but it obviates the need for the nuanced discussion of how an agent should act after entering a failure state when the mission is allowed to continue.} As a result, entering any failure state is considered to be an end of the mission, as in \cite{geibel2005}.  



The lifetime reward function $g: \mathcal{H}_{0:n} \rightarrow \mathbb{R}$ is defined as 
\begin{equation}
    g(h_{0:n}) = \sum_{t=0}^{n-1} \gamma^{t} R(s_{t},a_{t},s_{t+1}).
\end{equation}
To simplify notation, we introduce a binary random variable $C_{i}$ which is true if and only if the (stochastically determined) state $S_{i}$ is a safe state,
\begin{equation}
    C_{i} \iff S_{i} \in \mathcal{C}.
\end{equation}

Following the notation of \cite{santana2016}, we define the \emph{execution risk} of the policy $\pi$ following the state history $h_{0:t}$ as the probability that a failure state will be entered in the future after $h_{0:t}$ has occurred,
\begin{equation}
  er(h_{0:t},\pi) = \mathbb{P} \left[ \bigvee_{i=t+1}^{n} \overline{C}_{i} \middle| h_{0:t}, \pi \right].
\end{equation}
Unlike Santana et al., we do not consider partial observability of states, so in this work execution risk does not require consideration of the probability that any state in $h_{0:t}$ is a failure state, and only needs to be defined for safe state histories. When considering the state history $(s_{0})$, we will simply write $er(s_{0},\pi)$.


We seek the optimal policy $\pi^{*}$ that satisfies
\begin{equation}\label{eq:ProblemStatement}
\begin{array}{rcl}
     \pi^{*} = & \argmax_{\pi} & \mathbb{E}\left[g\left(H_{0:n} \right) \middle| s_{0}, \pi \right] \\
     & \text{s.t.} & er(s_{0},\pi) \leq \Delta \left( \mathbb{E}\left[ g(H_{0:n}) \middle| s_{0}, \pi \right] \right),
\end{array}
\end{equation}
for a specified concave nondecreasing function $\Delta: \mathbb{R} \rightarrow \left[0,1\right]$.

\section{Overview of Approach}

In general, methods that solve for the exact solution to problems similar to eq.~\ref{eq:ProblemStatement} do not scale well to large problems without heuristics, because the chance constraint couples all possible outcomes of the policy. An action on one branch may increase the expected reward, allowing more risk to be tolerated on a second branch with little relation to the first.

\begin{figure}[htb]
    \centering
    \includegraphics[width=12cm]{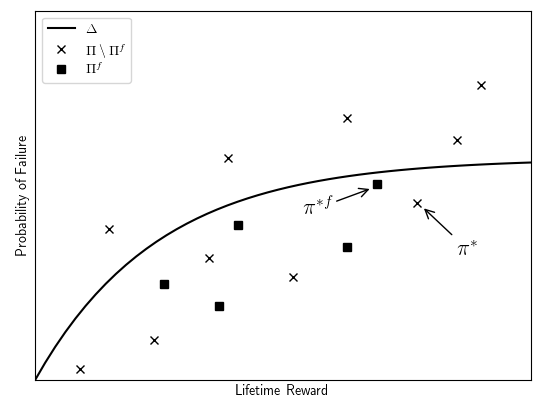}
    \caption{Policies visualized in terms of their lifetime reward and probability of failure. The constraint $\phi(f)$ defines a set of policies $\Pi^{f}$ that all lie below the line $\Delta$. The optimal solution is $\pi^{*}$, while Vulcan converges to the suboptimal policy $\pi^{*f}$.}
    \label{fig:Policy_Space}
\end{figure}

Our approach is instead to define a constraint $\phi$ which depends on a function $f$ and applies to state histories. If $\phi(f)$ is satisfied by every safe state history reachable by a policy, then that policy is guaranteed to satisfy the risk bounding function. Implicitly, the constraint defines a set of acceptable safe state histories $\mathcal{H}^{f}_{0:n} \subseteq \mathcal{H}^{\mathcal{C}}_{0:n}$, and the set of policies $\Pi^{f}$ that can be constructed from safe state histories in $\mathcal{H}^{f}_{0:n}$ all satisfy the chance constraint, so it need not be considered explicitly. The final solution returned is the highest reward policy in $\Pi^{f}$, which we call $\pi^{*f}$. Satisfaction of the constraint by all safe state histories is a sufficient but not necessary condition, so that a suboptimal solution $\pi^{*f}$ may be found, but optimality is traded off against an increased search speed. This idea is shown in figure~\ref{fig:Policy_Space} where $\Pi^{f}$ is a subset of all policies $\Pi$. All policies in $\Pi^{f}$ lie below the risk bounding function, and therefore satisfy the chance constraint, but there exist policies below the risk bounding function (potentially including the true optimal policy) that are not in $\Pi^{f}$. 

The advantage of this technique is that each of the constraints is \emph{local}, in the sense that the constraint can be verified for a state history without knowledge of the rest of the policy. The optimal policy in $\Pi^{f}$ can therefore be found using forward search and Bellman backups, like in an unconstrained MDP, and each state history only needs to be considered at most once. 

Once $f$ has been specified, search progresses by assuming that $\mathcal{H}^{f}_{0:n} = \mathcal{H}^{\mathcal{C}}_{0:n}$, meaning that every safe state sequence satisfies $\phi(f)$. Once a state history is explored up to the planning horizon, it is evaluated against $\phi(f)$. If $\phi(f)$ is satisfied, search continues without modification, but if $\phi(f)$ is violated or no actions remain from a state, then the preceding action is deleted. Search then continues as if the deleted action never existed. The end result is conceptually similar to search performed over $\mathcal{H}^{f}_{0:n}$, with additional state histories that are found and ignored when they are identified to not be part of $\mathcal{H}^{f}_{0:n}$. 

We proceed by reviewing related work. We then define the constraint $\phi(f)$ and introduce VulcanFS; which uses the constraint with forward search. VulcanFS is useful for smaller problems in that it still runs much faster than optimal explicit methods, and returns the policy $\pi^{*f}$ that the MCTS algorithm Vulcan will converge to. Since it searches all states, it also provides an upper bound for the run time of an MCTS based approach. We then use insights from the application of UCT \cite{kocsis2006} to MDPs in order to develop Vulcan; our MCTS based planner for CCMDPs.



\section{Related Work}

\subsection{Constrained MDPs and Chance Constrained MDPs}

Under our assumption of terminal failure states, a CCMDP may be formulated as a constrained MDP (CMDP)~\cite{altman1999}, which bounds the expected sum of costs gained from each action. In this case, a cost of 1 is received for entering a failure state, and 0 otherwise. All CMDP methods finding policies for states can also be made to return history-dependent policies by making a unique state for each state history. The major extension of our problem over a traditional CMDP is the introduction of a chance constraint that is a general concave nondecreasing function $\Delta(\cdot)$ as opposed to a constant, and that the state space is too large and expensive to compute in full.



%
%

Commonly used methods for finding policies for CCMDPs and CMDPs are linear programming and dynamic programming, while heuristic search and penalty methods also exist. Both linear programming and dynamic programming methods can be extended to handle certain classes of risk bounding functions, but their major drawback for our purposes is that they are unsuitable for the large state spaces we consider. Stochastic policies for CMDPs may be found in a time polynomial in the size of the CMDP~\cite{altman1999}, but it is known that computing deterministic policies for CMDPs is NP-complete~\cite{feinberg2000}. It follows that when the state spaces are too large to generate in full, neither linear programming nor dynamic programming methods are able to efficiently generate policies, and the problem is exacerbated when deterministic policies are desired. 

In unconstrained MDPs, a common approach is to use an online receding horizon approach to plan up to a reduced horizon $n'<n$, and repeating the process for outcomes of the previously found policy. The resulting policy is typically suboptimal, though it is possible to place guarantees on the suboptimality of the resulting policy \cite{alden1992,chang2003}. In CMDPs, receding horizon approaches can guarantee that risk bounds are satisfied over the planning horizon, but do not account for actions later in the policy without estimates for future risk, so receding horizon approaches give no guarantees on feasibility. Methods of using receding horizon planning for CMDPs such as \cite{undurti2010} require additional planning to guarantee feasibility from states, which may itself require searching through large state spaces and is additionally complicated when the allowed cost depends on the reward.

Our MCTS approach solves the problem of having to evaluate every state and reward in larger problems by running in an anytime manner, with the reward function computed only as states are explored. Application of UCT ensures that search is directed towards promising solutions so that a high scoring policy is achieved even when only a small fraction of the state space has been explored. While certain outcomes are not explored up to the full horizon $n$, exploration deep into the search tree is incorporated into the policy, and high probability outcomes are more likely to be considered. In other words, the returned policy is likely to satisfy the risk bounds up to the planning horizon for high probability outcomes, while in an incomplete policy the states that satisfy the risk bound only up to a reduced horizon are much less likely to actually occur.

\subsubsection{Linear Programming Methods}
Just as an MDP may be solved with a linear programming (LP) approach, there exists an LP formulation for CMDPs~\cite{heyman1982}. Deterministic policies can be computed instead using a mixed integer linear program (MILP) formulation, which adds a binary variable for each state-action pair in the CMDP~\cite{dolgov2005}. 

This method can be trivially extended to handle linear risk bounding functions, but general concave functions will require linear approximations. In theory the approximations can be made to reach an arbitrary degree of accuracy with increasing numbers of binary variables~\cite{bisschop2006}, but the more pressing issue is that the MILP becomes exponentially more difficult to solve with increasing numbers of binary variables. 



Furthermore, LP methods still require evaluation of every reward and cost function in the CMDP before a solution can be found. In our motivating example, when missions have a large number of measurements and measurement outcomes, performing this computation can be prohibitively expensive.

\subsubsection{Dynamic Programming Methods}
To use dynamic programming in a CMDP, a Lagrangian function is formulated which includes the CMDP reward and constraints. The problem of finding the optimal constrained policy then transforms into an unconstrained optimization over both the policy and the values of Lagrange multipliers in the Lagrangian function~\cite{altman1999}. For any given value of the Lagrange multipliers, the unconstrained optimal policy is found using standard dynamic programming. This approach has been applied for CCMDP problems with up to millions of states \cite{ono2012}.

Again, it is clear how to handle linear risk bounding functions, but general concave risk bounding functions cannot directly be used with dynamic programming. In addition, applying dynamic programming requires evaluation of all rewards and costs in the CMDP for each iteration of the algorithm used to select the Lagrange multipliers, which has the same drawbacks for large problems as the LP approach. By using MCTS, we accept a small degree of suboptimality in order to find policies for much larger chance constrained problems. 


\subsubsection{Heuristic Forward Search Methods}
An alternative approach to solving CCMDPs or CMDPs is the use of heuristic forward search similar to AO*. RAO*~\cite{santana2016} was designed for partially observable CCMDPs and uses both a reward heuristic and a risk heuristic in order to guide exploration of the search space. RAO* explicitly computes expected lifetime reward and execution risk, and so may be extended to be used with an arbitrary risk bounding function. While optimal deterministic policies are computed rapidly for very large problems when effective heuristics are available, it can be difficult to develop strong heuristics when the environment is not well known. In the worst case, RAO* must re-compute the optimal policy whenever new states are added to its search tree, which can lead to enumeration of the entire policy space. 

Vulcan does not require heuristics to identify high reward actions. When domain knowledge is present, it can be incorporated into the default policy used for expansion, while the degree of certainty in that default policy is controlled through UCT's exploration parameter. The default policy only needs to express an estimate for the best action to take at any state, which is more intuitive to specify than a numerical prediction of upcoming reward or risk. 

In the worst case, Vulcan will explore the entire state space before converging to its final policy, but it does not need to explore the entire policy space. In practice, the use of UCT causes rapid convergence towards high reward policies even without domain specific knowledge. 


\subsubsection{Risk Bounds as Penalties}

It has been proposed that chance constraints in CCMDPs, or more generally constraints in CMDPs, may be handled without reasoning over the constraints explicitly by adding a penalty to the reward function 
\begin{equation}
    \hat{R}(s_{t},a_{t},s_{t+1}) = R(s_{t},a_{t},s_{t+1}) - M \, \mathbb{P} \left[ \overline{C}_{t+1} \middle| s_{t}, a_{t} \right]
\end{equation}
and solving an unconstrained MDP for one or many constant values $M$. \cite{geibel2005} apply this approach for CCMDPs and use reinforcement learning on the resulting unconstrained MDP. Solving for the optimal policy is performed repeatedly, and the constant $M$ is reduced so long as computed policies are observed to obey a constant risk bound.

While solving an unconstrained MDP is easy compared to a CCMDP, it is not possible to guarantee that solutions to penalty methods satisfy risk bounds without explicitly computing the probability of failure of the policy found. Further, the best choice of $M$ depends on the values of the rewards, and there is no method to estimate a value of $M$ that leads to a satisfactory solution, which explains why repeated solutions with different values are necessary. In fact, \cite{undurti2010} showed that CCMDPs exist for which no value of $M$ gives the optimal policy. In contrast, our approach applies for all concave nondecreasing risk bounding functions, including constant functions, and even though our approach may also introduce some degree of suboptimality, the solution is guaranteed to satisfy the risk bounding function, and the CCMDP does not have to be repeatedly solved.

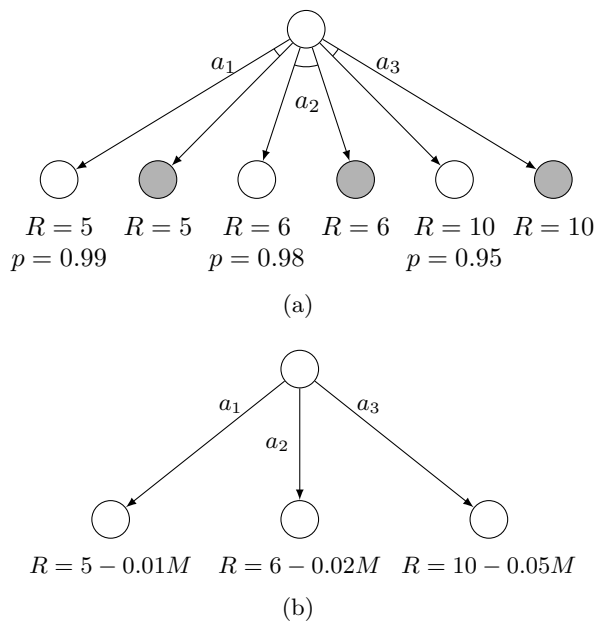
\begin{figure}
    \centering
    \subfloat[]{
    \begin{tikzpicture}[->,auto,>=latex,font=\small,node distance=0mm,sibling distance=8mm,level distance=20mm]
    \tikzset{edge from parent/.style={draw, edge from parent path={(\tikzparentnode) -- (\tikzchildnode)}}}
    
    \tikzstyle{round} = [thin,draw=black,circle,minimum size = 5mm]
    \tikzstyle{risk-round} = [thin,draw=black,circle,fill=black!30,minimum size = 5mm]
    \tikzstyle{invis} = [draw=none,minimum size = 0mm]
    
    \Tree[.\node[round] (s0) {};
           \edge node[pos=0.2,left,align=center]{$a_{1}$};
           \node[round] (s1) {};
           \node[risk-round] (s2) {};
           \node[round] (s3) {};
           \edge node[pos=0.5,left,align=center]{$a_{2}$};
           \node[risk-round] (s4) {};
           \node[round] (s5) {};
           \edge node[pos=0.2,right,align=center]{$a_{3}$};
           \node[risk-round] (s6) {};]
            
    \path pic[draw, -, angle radius=5mm] {angle = s1--s0--s2};
    \path pic[draw, -, angle radius=5mm] {angle = s3--s0--s4};
    \path pic[draw, -, angle radius=5mm] {angle = s5--s0--s6};
    
    \node[draw=none,rectangle,below=1mm of s1,align=center] {$R = 5$\\$p = 0.99$};
    \node[draw=none,rectangle,below=1mm of s2,align=center] {$R = 5$};
    \node[draw=none,rectangle,below=1mm of s3,align=center] {$R = 6$\\$p = 0.98$};
    \node[draw=none,rectangle,below=1mm of s4,align=center] {$R = 6$};
    \node[draw=none,rectangle,below=1mm of s5,align=center] {$R = 10$\\$p = 0.95$};
    \node[draw=none,rectangle,below=1mm of s6,align=center] {$R = 10$};
    
    \end{tikzpicture}}\\
    \subfloat[]{
    \begin{tikzpicture}[->,auto,>=latex,font=\footnotesize,node distance=0mm,sibling distance=20mm,level distance=20mm]
    \tikzset{edge from parent/.style={draw, edge from parent path={(\tikzparentnode) -- (\tikzchildnode)}}}
    
    \tikzstyle{round} = [thin,draw=black,circle,minimum size = 5mm]
    \tikzstyle{risk-round} = [thin,draw=black,circle,fill=black!30,minimum size = 5mm]
    \tikzstyle{invis} = [draw=none,minimum size = 0mm]
    
    \Tree[.\node[round] (s0) {};
           \edge node[pos=0.2,left,align=center]{$a_{1}$};
           \node[round] (s1) {};
           \edge node[pos=0.5,left,align=center]{$a_{2}$};
           \node[round] (s3) {};
           \edge node[pos=0.2,right,align=center]{$a_{3}$};
           \node[round] (s5) {};]
    
    \node[draw=none,rectangle,below=1mm of s1,align=center] {$R = 5-0.01M$};
    \node[draw=none,rectangle,below=1mm of s3,align=center] {$R = 6-0.02M$};
    \node[draw=none,rectangle,below=1mm of s5,align=center] {$R = 10-0.05M$};
    
    \end{tikzpicture}}
    
    \caption{An example of an MDP for which no penalty coefficient exists giving the optimal policy. (a) is the original MDP with failure states in gray, (b) is the same MDP encoded with probability of failure expressed as a penalty.}
    \label{fig:linear_penalty_failure}
\end{figure}

Even if we restrict our attention to certain classes of risk bounding functions such as linear functions, it is still not possible to guarantee that a value of $M$ that leads to the optimal solution exists, and so chance constraints may not be encoded as penalties in general. As an example, consider the very simple MDP with one action to be chosen in figure~\ref{fig:linear_penalty_failure} (a), with the objective of maximizing reward with a risk bounding function $\Delta(x) = 0.004 x$. By inspection, action $a_{3}$ exceeds the allowed risk bound, while action $a_{2}$ is the maximum reward choice that satisfies the risk bound. 

Formulating the risk as penalties on rewards leads to the MDP in figure~\ref{fig:linear_penalty_failure} (b), but for $M < 125$ the optimal action is $a_{3}$ and for $M > 125$ the optimal action is $a_{1}$, with a tie between the two at $M = 125$. Action $a_{2}$ is never selected, so a penalty method does not work here, while Vulcan will find the true optimal policy. 


\subsection{Application of MCTS to MDPs}

The application of Monte Carlo Tree Search to Markov Decision Processes has allowed approximately optimal policies to be found for previously intractably large MDPs, with perhaps the most visible application being the highly publicized success of AlphaGo \cite{silver2016}. MCTS methods produce a high reward policy without explicitly enumerating all states in the MDP. Instead, random samples are used to estimate the possible reward for an action from a state, and an action is only allowed additional samples so long as the results of previous samples compare favorably to other actions.

There are many variants of MCTS, based on different heuristics and selection rules \cite{browne2012}, but many build off the UCT algorithm introduced by~\cite{kocsis2006}. UCT balances \emph{exploration} of actions that have not been frequently sampled against \emph{exploitation} of actions that previous samples have implied return a large reward. To do so, UCT dictates that all actions should be sampled from a state once, and then subsequent sample actions should be selected according to
\begin{equation}\label{eq:uct_action}
    a_{uct} = \argmax_{a} \ \Tilde{Q}(s,a) + c \ \sqrt{\frac{\log N_{s}}{N_{s,a}}}
\end{equation}
\begin{equation}
    \Tilde{Q}(s_{t},a) = \sum_{s_{t+1}} \frac{N_{s_{t+1}}}{N_{s,a}} \left( R(s_{t},a,s_{t+1}) + \max_{a'} \gamma \ \Tilde{Q}(s_{t+1},a') \right).
\end{equation}
Where $N_{s}$ is the total number of samples performed from state $s$, $N_{s,a}$ is the total number of samples of action $a$ taken from state $s$, $c$ is a tuned constant, and $\Tilde{Q}(s,a)$ is a numerical estimate for the action value function (reward to go) from $s$ based on the previous samples taken during tree search. An action is more likely to be chosen by UCT if it has a high action value function, or the number of times it has been sampled is low compared to the total number of samples taken. After choosing a sample action, the specific outcome to be sampled next is selected randomly according to the probability distribution of outcomes. 

The significance of UCT is that for certain choices of $c$, the expected error in $\Tilde{Q}$ shrinks at a rate $\mathcal{O} \left( \log N / N \right)$ and the probability of finding a suboptimal policy can be shown to decay to zero at a polynomial rate~\cite{kocsis2006}. In practice $c$ is often tuned empirically, though a theoretical value of $\sqrt{2}$ can be derived for the bandit setting~\cite{auer2002}.

While UCT works well for large MDPs, it has not been applied to constrained or chance constrained MDPs because of coupling between different outcomes through the constraint, which means that the optimal policy does not maximize a value function at a single state. By decoupling the constraint, Vulcan allows UCT to be applied to a CCMDP, gaining fast convergence to an approximately optimal policy.

%

\section{Local Constraints for Risk Bounding}

In this section we introduce and prove our local constraints on safe state histories which guarantee that the risk bounding function is satisfied. These constraints form the basis our forward search and MCTS based algorithms. The constraints involve a property of state histories that we call the \emph{sequence execution risk}, based off of the previously defined execution risk.

\subsection{Definition of Sequence Execution Risk}

The sequence execution risk of a state history $h_{t:n}$ is related to the combined probability of entering a failure state by executing $a_{t}$ from $s_{t}$, $a_{t+1}$ from $s_{t+1}$, and so forth until $a_{n-1}$ from $s_{n-1}$, assuming all states in the history are safe. For state histories with failure states, the $ser$ is zero by definition. Formally,
\begin{equation}\label{eq:SerDefinition}
    ser(h_{t:n}) = 
      \frac{\mathbb{P}\left[ \bigvee_{i=t}^{n-1} \left( \overline{C}_{i+1} \middle| s_{i}, a_{i} \right) \right]}{1 - \mathbb{P}\left[ \bigvee_{i=t}^{n-1} \left( \overline{C}_{i+1} \middle| s_{i}, a_{i} \right) \right]} \mathbf{1}_{\mathcal{H}_{t:n}^{\mathcal{C}}}(h_{t:n}),
\end{equation}
where $\mathbf{1}_{\mathcal{H}_{t:n}^{\mathcal{C}}}(h_{t:n})$ is the indicator function that is one if $h_{t:n} \in \mathcal{H}_{t:n}^{\mathcal{C}}$ and zero otherwise.

We using the following shorthand for the immediate probability of failure of taking action $a_{t}$ from state $s_{t}$, which may be thought of as the immediate risk of the action:
\begin{equation}
    r(s_{t},a_{t}) = \mathbb{P} \left[ \overline{C}_{t+1} \middle| s_{t}, a_{t} \right].
\end{equation}

Then the sequence execution risk may be efficiently computed (or alternatively defined) for safe state histories as:
\begin{equation}
    ser(h_{t:n}) = \frac{1 - \prod_{i=t}^{n-1} \left( 1 - r(s_{i},a_{i}) \right)}{\prod_{i=t}^{n-1} \left( 1 - r(s_{i},a_{i})\right)} \mathbf{1}_{\mathcal{H}_{t:n}^{\mathcal{C}}}(h_{t:n}).
\end{equation}

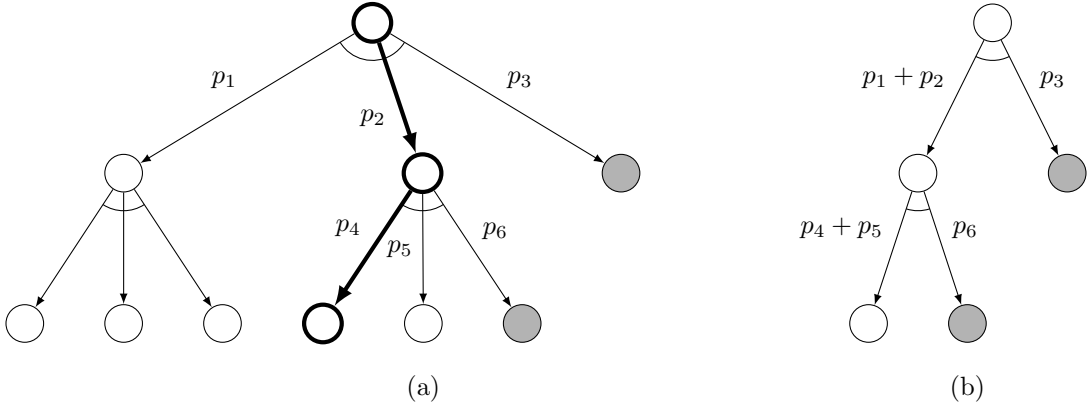
\begin{figure}
    \centering
    \begin{tikzpicture}[->,auto,>=latex,font=\small,sibling distance=8mm,level distance=20mm,level 1/.style={level distance = 0mm,sibling distance = 20mm}]
    \tikzset{edge from parent/.style={draw, edge from parent path={(\tikzparentnode) -- (\tikzchildnode)}}}
    
    \tikzstyle{round} = [thin,draw=black,circle,minimum size = 5mm]
    \tikzstyle{major-round} = [ultra thick,draw=black,circle,minimum size = 5mm]
    \tikzstyle{risk-round} = [thin,draw=black,circle,fill=black!30,minimum size = 5mm]
    \tikzstyle{minor-round} = [thin,draw=black!20,circle,text=black!20,minimum size = 5mm]
    
    \Tree [\edge [draw=none];
           [.\node[major-round] (s0) {}; 
             \edge node[auto=right]{$p_{1}$};
             [.\node[round] (s1) {};
              \node[round] (s4) {};
              \node[round] (s5) {};
              \node[round] (s6) {};]
             \edge [ultra thick] node[auto=right]{$p_{2}$};
             [.\node[major-round] (s2) {};
              \edge [ultra thick] node[auto=right]{$p_{4}$};
              \node[major-round] (s7) {};
              \edge node[auto=right]{$p_{5}$};
              \node[round] (s8) {};
              \edge node[auto=left]{$p_{6}$};
              \node[risk-round] (s9) {};]
             \edge node[auto=left]{$p_{3}$};
             \node[risk-round] (s3) {};]
           \edge [draw=none];
           [.\node[round] (s0-ser) {};
             \edge node[auto=right]{$p_{1} + p_{2}$};
             [.\node[round] (s1-ser) {};
              \edge node[auto=right]{$p_{4} + p_{5}$};
              \node[round] (s3-ser) {};
              \edge node[auto=left]{$p_{6}$};
              \node[risk-round] (s4-ser) {};]
            \edge node[auto=left]{$p_{3}$};
            \node[risk-round] (s2-ser) {};]]
            
    \path pic[draw, -, angle radius=5mm] {angle = s1--s0--s3};
    \path pic[draw, -, angle radius=5mm] {angle = s4--s1--s6};
    \path pic[draw, -, angle radius=5mm] {angle = s7--s2--s9};
    
    \path pic[draw, -, angle radius=5mm] {angle = s1-ser--s0-ser--s2-ser};
    \path pic[draw, -, angle radius=5mm] {angle = s3-ser--s1-ser--s4-ser};
    
    \node[draw=none,rectangle,below=3mm of s8] {(a)};
    \node[draw=none,rectangle,below=3mm of s4-ser] {(b)};
    
    \end{tikzpicture}
    \caption{Visual interpretation of the numerator of the sequence execution risk of the state history highlighted in (a) as the execution risk of the only policy possible in the MDP in (b). Probability of failure remains the same, and remaining probability is collapsed into the safe state in the history. Gray states indicate failure states.}
    \label{fig:visual-ser}
\end{figure}

Intuitively, the numerator of the $ser$ of a state history may be interpreted as the execution risk of the only policy possible in a different MDP, where only the actions taken in the state history are available, and taking the action $a_{t}$ from state $s_{t}$ leads to a failure state with probability $\mathbb{P} \left[ \overline{C}_{t+1} \middle| s_{t}, a_{t} \right]$ (the same probability as in the original policy) and state $s_{t+1}$ otherwise. This intuitive definition is visualized in figure~\ref{fig:visual-ser}. The denominator acts as a scaling factor to give sequence execution risk a desirable expectation.

\subsection{Local Constraint and Proof}

Using the sequence execution risk, we now introduce our constraint $\phi(f)$ that ensures a policy satisfies the risk bounding function. Let $f: \mathcal{H}_{0:n} \rightarrow \mathbb{R}$ be any function with expectation bounded by the expected cumulative reward in the policy:
\begin{equation}\label{eq:fCondition}
    \mathbb{E}\left[ f(H_{0:n}) \middle| s_{0}, \pi \right] \leq \mathbb{E}\left[ g(H_{0:n}) \middle| s_{0}, \pi \right].
\end{equation}

The following theorem shows that a local constraint on safe state histories that ensures the risk bounding function is satisfied can be created for every such $f$. 

\begin{theorem}\label{th:MajorTheorem}
For a concave nondecreasing function $\Delta$, asserting that for all state histories $h_{0:n}$ reachable under policy $\pi$
\begin{equation*}
    ser(h_{0:n}) \leq \Delta \left( f(h_{0:n}) \right)
\end{equation*}
implies
\begin{equation*}
    er(s_{0},\pi) \leq \Delta \left( \mathbb{E}\left[ g(H_{0:n}) \middle| s_{0}, \pi \right] \right).
\end{equation*}
\end{theorem}

Here, the constraint $ser(h_{0:n}) \leq \Delta \left( f(h_{0:n}) \right)$ is our previously mentioned $\phi(f)$. Since $ser$ is zero for unsafe state histories and $\Delta$ maps to $[0,1]$, the condition of theorem~\ref{th:MajorTheorem} is automatically satisfied for all unsafe state histories.

The implication of satisfying $\phi(f)$ for all state histories is exactly the constraint on our problem statement in eq.~\ref{eq:ProblemStatement}, so a policy may be found by choosing an appropriate function $f$, and ensuring that $\phi(f)$ is satisfied for each safe state history. The basis of our algorithms is to search through feasible safe state histories, evaluate $ser$ and $f$, and construct the highest scoring policy from among those state histories that satisfy the local constraint. 

To prove theorem~\ref{th:MajorTheorem}, we begin with two lemmas.

\begin{lemma}\label{lem:MinorLemma}
\begin{equation*}
    \sum_{h_{t:n} \in \mathcal{H}_{t:n}^{\mathcal{C}}} \frac{p(h_{t:n}|h_{0:t},\pi)}{\prod_{i=t}^{n-1} \left(1 - r(s_{i},a_{i}) \right)} = 1,
\end{equation*}
%
where $s_{i}$, $a_{i}$ are the elements of the state history $h_{t:n}$ and $a_{i}=\pi(h_{0:i})$.
\end{lemma}

\begin{proof}
Our proof is by induction. For the case $t=n-1$, we have
%
\begin{equation}
    \sum_{\mathclap{h_{n-1:n} \in \mathcal{H}_{n-1:n}^{\mathcal{C}}}} p(h_{n-1:n}|h_{0:n-1},\pi) = \sum_{s_{n} \in \mathcal{C}} p(s_{n}|h_{0:n-1},\pi) = 1 - r(s_{n-1},a_{n-1}),
\end{equation}
so the lemma holds.

Assuming the lemma is true for $t+1$, the case of $t$ is proven as
\begin{equation}
\begin{split}
    \sum_{h_{t:n} \in \mathcal{H}_{t:n}^{\mathcal{C}}} \frac{p(h_{t:n}|h_{0:t},\pi)}{\prod_{i=t}^{n-1} \left(1 - r(s_{i},a_{i}) \right)} &= \sum_{s_{t+1} \in \mathcal{C}} \frac{p(s_{t+1}|h_{0:t},\pi)}{1-r(s_{t},a_{t})} \ \sum_{\mathclap{h_{t+1:n} \in \mathcal{H}_{t+1:n}^{\mathcal{C}}}}  \quad \frac{p(h_{t+1:n}|h_{0:t+1},\pi)}{\prod_{i=t+1}^{n-1} \left( 1 - r(s_{i},a_{i}) \right)} \\
    &= \sum_{s_{t+1} \in \mathcal{C}} \frac{p(s_{t+1}|h_{0:t},\pi)}{1-r(s_{t},a_{t})} \\
    &= 1.
\end{split}
\end{equation}
\end{proof}

\begin{lemma}\label{lem:MajorLemma}
\begin{equation*}
    er(h_{0:t},\pi) = \mathbb{E} \left[ ser\left( H_{t:n} \right) \middle| h_{0:t}, \pi \right],
\end{equation*}
where the expectation with the conditions is understood to be over all state histories of the form $(s_{t},\pi(h_{0:t}),...)$.
\end{lemma}

\begin{proof}
We note the following two recursive expressions for execution risk and sequence execution risk:
\begin{equation}
    er(h_{0:n},\pi) = 0,
\end{equation}
\begin{equation}
    er(h_{0:t},\pi) = r(s_{t},a_{t}) + \sum_{s_{t+1} \in \mathcal{C}} p(s_{t+1}| h_{0:t}, \pi) \ er(h_{0:t+1},\pi),
\end{equation}
where $a_{t} = \pi(h_{0:t})$, and for safe state histories,
\begin{equation}
    ser(h_{n}) = 0,
\end{equation}
\begin{equation}
\begin{split}
    ser(h_{t:n}) &= \frac{r(s_{t},a_{t}) + \left( 1-r(s_{t},a_{t}) \right) - \prod_{i=t}^{n-1} \left(1 - r(s_{i},a_{i}) \right)}{\prod_{i=t}^{n-1} \left( 1 - r(s_{i},a_{i}) \right)} \\
    &= \frac{r(s_{t},a_{t})}{\prod_{i=t}^{n-1} \left( 1 - r(s_{i},a_{i}) \right)} + ser(h_{t+1:n}).
\end{split}
\end{equation}

%

The proof then follows from induction. The case where $t=n$ is trivially true since $er(h_{0:n},\pi) = ser(h_{n}) = 0$. Assuming the lemma is true for $t+1$, the case of $t$ follows from
\begin{equation}\label{eq:LemmaProof1}
\begin{split}
    er(h_{0:t},\pi) &= r(s_{t},a_{t}) + \sum_{s_{t+1} \in \mathcal{C}} p(s_{t+1}|h_{0:t}, \pi) \ er(h_{0:t+1},\pi) \\
    &= r(s_{t},a_{t}) + \sum_{s_{t+1} \in \mathcal{C}} p(s_{t+1}|h_{0:t}, \pi) \ \mathbb{E} \left[ ser(H_{t+1:n}) \middle| h_{0:t+1}, \pi \right] \\
    &= r(s_{t},a_{t}) + \sum_{s_{t+1} \in \mathcal{C}} p(s_{t+1}|h_{0:t}, \pi) \sum_{\mathclap{h_{t+1:n} \in \mathcal{H}_{t+1:n}^{\mathcal{C}}}} p\left( h_{t+1:n} \middle| h_{0:t+1}, \pi \right) ser(h_{t+1:n}) \\
    &= r(s_{t},a_{t}) + \sum_{\mathclap{h_{t:n} \in \mathcal{H}_{t:n}^{\mathcal{C}}}} p\left( h_{t:n} \middle| h_{0:t}, \pi \right) ser(h_{t+1:n}) \\
    &= \sum_{\mathclap{h_{t:n} \in \mathcal{H}_{t:n}^{\mathcal{C}}}} p(h_{t:n}|h_{0:t},\pi) \left( \frac{r(s_{t},a_{t})}{\prod_{i=t}^{n-1} \left( 1 - r(s_{i},a_{i}) \right)} + ser(h_{t+1:n}) \right) \\
    &= \sum_{\mathclap{h_{t:n} \in \mathcal{H}_{t:n}^{\mathcal{C}}}} p(h_{t:n}|h_{0:t},\pi) ser(h_{t:n}) \\
    &= \mathbb{E} \left[ ser(H_{t:n}) | h_{0:t}, \pi \right].
\end{split}
\end{equation}
Since $ser$ is defined to be zero for state histories involving failure states, the expectation needs only to sum over safe state histories. The additional terms from moving $r(s_{t},a_{t})$ into the sum follow from lemma~\ref{lem:MinorLemma}.

\end{proof}

Using lemma~\ref{lem:MajorLemma}, Theorem~\ref{th:MajorTheorem} follows simply.

\begin{proof}[Proof of Theorem~\ref{th:MajorTheorem}]
\begin{equation}
\begin{split}
    er(s_{0},\pi) &= \mathbb{E} \left[ ser(H_{0:n}) \middle| s_{0}, \pi \right] \\
                  &\leq \mathbb{E}\left[ \Delta( f(H_{0:n}) ) \middle| s_{0}, \pi \right] \\
                  &\leq \Delta\left( \mathbb{E}\left[ f(H_{0:n}) \middle| s_{0}, \pi \right] \right) \\
                  &\leq \Delta\left( \mathbb{E}\left[ g(H_{0:n}) \middle| s_{0}, \pi \right] \right),
\end{split}
\end{equation}
where the third line follows from Jensen's inequality and the concavity of $\Delta$, while the fourth line follows from the facts that $\mathbb{E}\left[ f(H_{0:n}) \middle| s_{0}, \pi \right] \leq \mathbb{E}\left[ g(H_{0:n}) \middle| s_{0}, \pi \right]$ and $\Delta$ is non-decreasing.
\end{proof}

Note that since we have defined the sequence execution risk to be zero for any state history involving a failure state, $ser(h_{0:n}) \leq \Delta(f(h_{0:n}))$ is always true for any failing state history. This was done so that policies which take high risks up front in order to reach large rewards at later time steps are not forbidden due to large probabilities of failure but low rewards on short state histories that end in early failures. The price that is paid is an inflation of $ser$ for safe state histories so that it remains the expectation of the execution risk, which accounts for the denominator in eq.~\ref{eq:SerDefinition}.

\subsection{Selection of Function $f$}

We have defined a constraint which uses a function $f$. A choice of function for which eq.~\ref{eq:fCondition} is satisfied with equality will introduce less conservatism to the solution, but there are a large number of functions with this property, and it remains to be shown why it would be advantageous to choose a function other than the obvious $f = g$.  

\begin{figure}
    \centering
    
    \begin{tikzpicture}[->,auto,>=latex,font=\small,node distance=0mm,sibling distance=30mm,level distance=40mm,level 1/.style={level distance = 10mm}]
    \tikzset{edge from parent/.style={draw, edge from parent path={(\tikzparentnode) -- (\tikzchildnode)}}}
    
    \tikzstyle{round} = [thin,draw=black,circle,minimum size = 7mm]
    \tikzstyle{major-round} = [ultra thick,draw=black,circle,minimum size = 5mm]
    \tikzstyle{risk-round} = [thin,draw=black,circle,fill=black!30,minimum size = 5mm]
    \tikzstyle{minor-round} = [thin,draw=black!20,circle,text=black!20,minimum size = 5mm]
    
    \Tree[\edge [dotted,thick];
           [.\node[round] (s0) {};
             \edge node[auto=right,align=center]{$p$ high\\$g(h_{0:n})$ high};
             \node[round] (s1) {};
             \edge node[pos=0.8,left,align=center]{$p$ high\\$g(h_{0:n})$ high};
             \node[round] (s2) {};
             \edge node[auto=left,align=center]{$p$ low\\$g(h_{0:n})$ low};
             \node[round] (s3) {};]]
            
    \path pic[draw, -, angle radius=5mm] {angle = s1--s0--s3};
    
    \end{tikzpicture}

    \caption{Illustrative case where an unlikely low scoring outcome causes an action to be deleted in Vulcan.}
    \label{fig:Bad_f}
\end{figure}
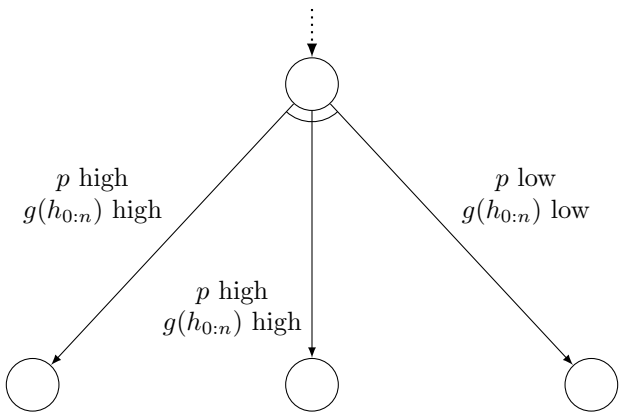

The downside of using $f=g$ is that unlikely and low reward outcomes of an action can lead to state histories with $ser(h_{0:n}) > \Delta(g(h_{0:n}))$, as in figure~\ref{fig:Bad_f}. A very low reward outcome with a low probability does not heavily influence the expected lifetime reward of a policy taking an action, but the state history including that outcome will have a much lower lifetime reward, and may not satisfy the risk bound, which prevents the action from being taken. This situation occurs frequently in CCMDPs with low probability of failure and zero reward received upon failure.

Instead, we recommend a function that averages reward from the outcomes of each action,
\begin{equation}
    f_{1}(h_{0:n}) = \sum_{t=0}^{n-1} \gamma^{t} \ \mathbb{E}\left[ R(s_{t}, a_{t}, S_{t+1}) \right].
\end{equation}
This function satisfies eq.~\ref{eq:fCondition} with equality, and since it averages the outcomes of actions, it avoids the case where unlikely low reward outcomes prevent an action from being included in the policy. The downside is that the rewards to multiple outcome states must be evaluated, which can be computationally expensive. Nonetheless, when used in our MCTS algorithm, the advantage of not needing to enumerate all states and reward functions remains. Our experimental results all use this function.

\section{VulcanFS: A Forward Search Based Algorithm}

Using theorem~\ref{th:MajorTheorem}, it is possible to derive a version of forward search \cite{Sucar2011} that ensures the chance constraint is satisfied, which we call VulcanFS. The algorithm is presented in algorithm~\ref{alg:ForwardSearch}. At the planning horizon, if $\phi(f)$ is violated, the reward is set to $-\infty$ to ensure the action is never taken. Otherwise, the lifetime reward is maximized. 

\begin{algorithm}\label{alg:ForwardSearch}
\caption{VulcanFS}
\KwIn{History to current state $h_{0:t}$.}
\KwOut{Value function $Q(h_{0:t},\pi)$. Sets policy to $\pi = \pi^{*f}$.}
\If{$s_{t}$ \textup{is a failure state}}
{\Return $Q(h_{0:t},\pi) = 0$}
\If{$t = n$}
{\eIf{$ser(h_{0:n}) \leq \Delta(f(h_{0:n}))$}
     {\Return $Q(h_{0:n},\pi) = 0$}
     {\Return $Q(h_{0:n},\pi) = -\infty$}}
\For{$a \in \mathcal{A}$}
    {$Q(h_{0:t},a) \leftarrow \sum\limits_{\mathclap{s_{t+1}}} p(s_{t+1}|h_{0:t},a) \left[ R(s_{t},a,s_{t+1}) + \gamma \ \textup{VulcanFS}(h_{0:t+1}) \right]$ \\}
    $\pi(h_{0:t}) \leftarrow \argmax\limits_{a} Q(h_{0:t},a)$ \\
    \Return $Q(h_{0:t},\pi) = \max\limits_{a} Q(h_{0:t},a)$
\end{algorithm}

Like forward search, the worst case complexity of algorithm~\ref{alg:ForwardSearch} is $\mathcal{O}((\left\vert \mathcal{A} \right\vert B)^{n})$ \cite{Sucar2011}, where $B$ is the maximum number of outcomes of a single action. This is much smaller than the number of possible policies $\mathcal{O}(\left\vert \mathcal{A} \right\vert^{B^{n-1}})$.

\section{Vulcan: An MCTS Based Algorithm}

Algorithm~\ref{alg:ForwardSearch} shows how a deterministic policy satisfying a risk-bounding function may be found by only considering each possible state history once. For problems where the rewards and probabilities of failure are functions of the entire state history and ordering, this corresponds to the state space of the problem. However, algorithm~\ref{alg:ForwardSearch} is still not suitable for large problems where evaluating every reward is not feasible; its main value comes from how it may be extended to the MCTS implementation Vulcan, given in algorithm~\ref{alg:Vulcan}. The idea behind Vulcan is to use the insights of UCT in order to construct a policy that rapidly converges towards $\pi^{*f}$ while only sampling a subset of state histories allowed by the MDP.

Theorem~\ref{th:MajorTheorem} is used to ensure that the policy does not violate the chance constraint. Naturally, if the algorithm is terminated before an entire policy has been evaluated down to the planning horizon, guarantees cannot be placed on the states that have not been examined. However, in this case theorem~\ref{th:MajorTheorem} asserts the chance constraint over the states that have been explored; bounding the probability of failure across those states as a function of lifetime reward computed across those states. 


Vulcan uses UCT to guide sampling in algorithm~\ref{alg:Sample}, which balances exploration against exploitation based on the reward and numbers of samples of state histories. Like in a traditional UCT algorithm, a default policy is used to incorporate domain-specific knowledge if it is available. However, our algorithm differs from standard UCT in three aspects.

First, since we are interested in domains where computation of reward is expensive, we avoid recomputing reward between states that have already been sampled. This is achieved by keeping all states visited in the search tree. Instead of expanding the search tree by a single node and performing a rollout, which evaluates many states but keeps only their cumulative reward, we add states to the search tree until reaching the planning horizon, saving the rollout in memory. This trades off memory usage against the slowdown of repeatedly evaluating an expensive state.

Second, we impose the constraint of theorem~\ref{th:MajorTheorem} on all safe state histories at the planning horizon. If the sequence execution risk does not satisfy $\phi(f)$, then the last action in the sampled state history is deleted so that it is not sampled again. Likewise, if all actions have been deleted from a non-terminal state history, then then taking any action from the sampled state history has a non-zero probability of leading to a state that violates the conditions of theorem~\ref{th:MajorTheorem}, so the last action is again removed. After an action is deleted, sampling continues from the parent node of that action, until a satisfactory state history has been found. Sample counts are only incremented when a state history is found that satisfies the enforced condition, and they are decremented when states are deleted. This way, the regret bounds derived for UCT apply, with regard to the number of samples taken from state histories in $\mathcal{H}^{f}$.

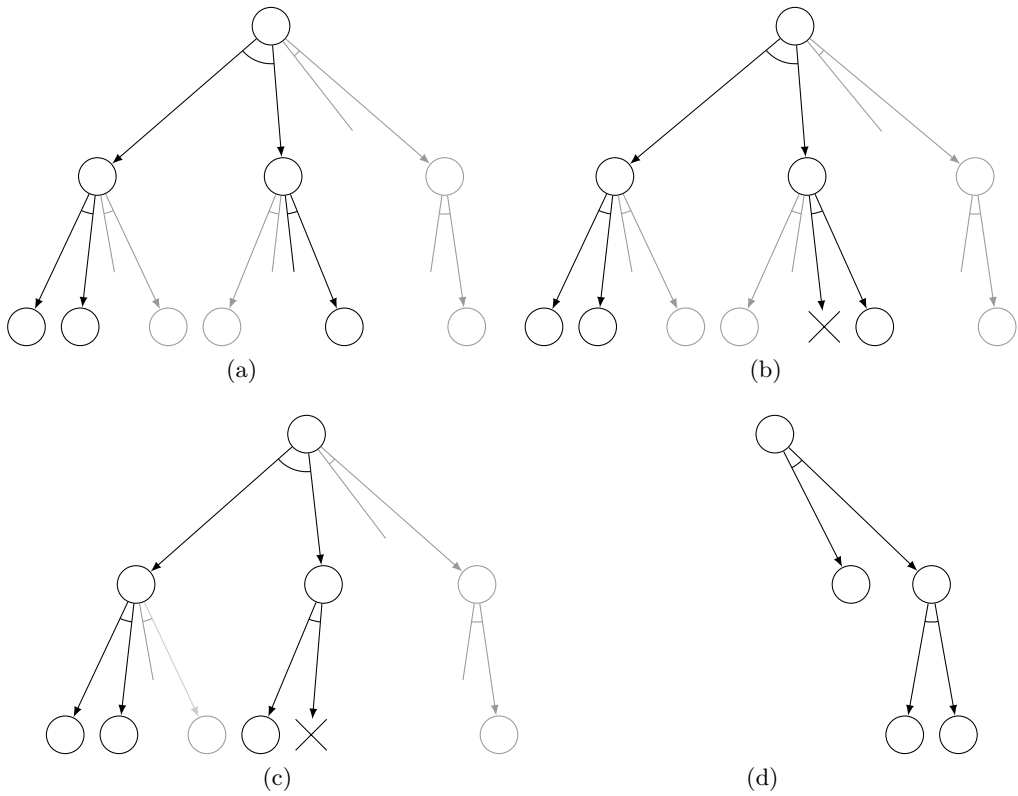
\begin{figure}
    \centering
    \subfloat[]{
    \begin{tikzpicture}[->,auto,>=latex,font=\small,node distance=0mm,sibling distance=2mm,level distance=20mm]
    \tikzset{edge from parent/.style={draw, edge from parent path={(\tikzparentnode) -- (\tikzchildnode)}}}
    
    \tikzstyle{round} = [thin,draw=black,circle,minimum size = 5mm]
    \tikzstyle{major-round} = [ultra thick,draw=black,circle,minimum size = 5mm]
    \tikzstyle{risk-round} = [thin,draw=black,circle,fill=black!30,minimum size = 5mm]
    \tikzstyle{minor-round} = [thin,draw=black!40,circle,text=black!20,minimum size = 5mm]
    \tikzstyle{invis} = [draw=none,minimum size = 0mm]
    
    \Tree[.\node[round] (s0) {};
             [.\node[round] (s1) {};
              \node[round] (s5) {};
              \node[round] (s6) {};
              \edge[-,draw=black!40,shorten >= 6mm];
              \node[invis] (s7) {};
              \edge[draw=black!40];
              \node[minor-round] (s8) {};]
             [.\node[round] (s2) {};
              \edge[draw=black!40];
              \node[minor-round] (s9) {};
              \edge[-,draw=black!40,shorten >= 6mm];
              \node[invis] (s10) {};
              \edge[-,shorten >= 6mm];
              \node[invis] (s11) {};
              \node[round] (s12) {};]
             \edge[-,draw=black!40,shorten >= 6mm];
             \node[invis] (s3) {};
             \edge[draw=black!40];
             [.\node[minor-round] (s4) {};
              \edge[-,draw=black!40,shorten >= 6mm];
              \node[invis] (s13) {};
              \edge[draw=black!40];
              \node[minor-round] (s14) {};]]
            
    \path pic[draw, -, angle radius=5mm] {angle = s1--s0--s2};
    \path pic[draw, -, angle radius=5mm] {angle = s5--s1--s6};
    \path pic[draw, -, angle radius=5mm, draw=black!40] {angle = s7--s1--s8};
    \path pic[draw, -, angle radius=5mm, draw=black!40] {angle = s9--s2--s10};
    \path pic[draw, -, angle radius=5mm] {angle = s11--s2--s12};
    \path pic[draw, -, angle radius=5mm, draw=black!40] {angle = s3--s0--s4};
    \path pic[draw, -, angle radius=5mm, draw=black!40] {angle = s13--s4--s14};
    
    \end{tikzpicture}}%
    \quad
    \subfloat[]{
    \begin{tikzpicture}[->,auto,>=latex,font=\footnotesize,node distance=0mm,sibling distance=2mm,level distance=20mm]
    \tikzset{edge from parent/.style={draw, edge from parent path={(\tikzparentnode) -- (\tikzchildnode)}}}
    
    \tikzstyle{round} = [thin,draw=black,circle,minimum size = 5mm]
    \tikzstyle{major-round} = [ultra thick,draw=black,circle,minimum size = 5mm]
    \tikzstyle{risk} = [cross out,draw,minimum size = 4mm]
    \tikzstyle{minor-round} = [thin,draw=black!40,circle,text=black!20,minimum size = 5mm]
    \tikzstyle{invis} = [draw=none,minimum size = 0mm]
    
    \Tree[.\node[round] (s0) {};
             [.\node[round] (s1) {};
              \node[round] (s5) {};
              \node[round] (s6) {};
              \edge[-,draw=black!40,shorten >= 6mm];
              \node[invis] (s7) {};
              \edge[draw=black!40];
              \node[minor-round] (s8) {};]
             [.\node[round] (s2) {};
              \edge[draw=black!40];
              \node[minor-round] (s9) {};
              \edge[-,draw=black!40,shorten >= 6mm];
              \node[invis] (s10) {};
              \node[risk] (s11) {};
              \node[round] (s12) {};]
             \edge[-,draw=black!40,shorten >= 6mm];
             \node[invis] (s3) {};
             \edge[draw=black!40];
             [.\node[minor-round] (s4) {};
              \edge[-,draw=black!40,shorten >= 6mm];
              \node[invis] (s13) {};
              \edge[draw=black!40];
              \node[minor-round] (s14) {};]]
            
    \path pic[draw, -, angle radius=5mm] {angle = s1--s0--s2};
    \path pic[draw, -, angle radius=5mm] {angle = s5--s1--s6};
    \path pic[draw, -, angle radius=5mm, draw=black!40] {angle = s7--s1--s8};
    \path pic[draw, -, angle radius=5mm, draw=black!40] {angle = s9--s2--s10};
    \path pic[draw, -, angle radius=5mm] {angle = s11--s2--s12};
    \path pic[draw, -, angle radius=5mm, draw=black!40] {angle = s3--s0--s4};
    \path pic[draw, -, angle radius=5mm, draw=black!40] {angle = s13--s4--s14};
    
    \end{tikzpicture}}
    
    \subfloat[]{
    \begin{tikzpicture}[->,auto,>=latex,font=\footnotesize,node distance=0mm,sibling distance=2mm,level distance=20mm]
    \tikzset{edge from parent/.style={draw, edge from parent path={(\tikzparentnode) -- (\tikzchildnode)}}}
    
    \tikzstyle{round} = [thin,draw=black,circle,minimum size = 5mm]
    \tikzstyle{major-round} = [ultra thick,draw=black,circle,minimum size = 5mm]
    \tikzstyle{risk} = [cross out,draw,minimum size = 4mm]
    \tikzstyle{minor-round} = [thin,draw=black!40,circle,text=black!20,minimum size = 5mm]
    \tikzstyle{invis} = [draw=none,minimum size = 0mm]
    
    \Tree[.\node[round] (s0) {};
             [.\node[round] (s1) {};
              \node[round] (s5) {};
              \node[round] (s6) {};
              \edge[-,draw=black!40,shorten >= 6mm];
              \node[invis] (s7) {};
              \edge[draw=black!20];
              \node[minor-round] (s8) {};]
             [.\node[round] (s2) {};
              \node[round] (s9) {};
              \node[risk] (s10) {};
              \edge[draw=none];
              \node[invis] (s11) {};
              \edge[draw=none];
              \node[invis] (s12) {};]
             \edge[-,draw=black!40,shorten >= 6mm];
             \node[invis] (s3) {};
             \edge[draw=black!40];
             [.\node[minor-round] (s4) {};
              \edge[-,draw=black!40,shorten >= 6mm];
              \node[invis] (s13) {};
              \edge[draw=black!40];
              \node[minor-round] (s14) {};]]
            
    \path pic[draw, -, angle radius=5mm] {angle = s1--s0--s2};
    \path pic[draw, -, angle radius=5mm] {angle = s5--s1--s6};
    \path pic[draw, -, angle radius=5mm, draw=black!40] {angle = s7--s1--s8};
    \path pic[draw, -, angle radius=5mm] {angle = s9--s2--s10};
    \path pic[draw, -, angle radius=5mm, draw=black!40] {angle = s3--s0--s4};
    \path pic[draw, -, angle radius=5mm, draw=black!40] {angle = s13--s4--s14};
    
    \end{tikzpicture}}%
    \quad
    \subfloat[]{
    \begin{tikzpicture}[->,auto,>=latex,font=\footnotesize,node distance=0mm,sibling distance=2mm,level distance=20mm]
    \tikzset{edge from parent/.style={draw, edge from parent path={(\tikzparentnode) -- (\tikzchildnode)}}}
    
    \tikzstyle{round} = [thin,draw=black,circle,minimum size = 5mm]
    \tikzstyle{major-round} = [ultra thick,draw=black,circle,minimum size = 5mm]
    \tikzstyle{risk} = [cross out,draw,minimum size = 4mm]
    \tikzstyle{minor-round} = [thin,draw=black!20,circle,text=black!20,minimum size = 5mm]
    \tikzstyle{invis} = [draw=none,minimum size = 0mm]
    
    \Tree[.\node[round] (s0) {};
          \edge[draw=none];
             [.\node[invis] (s1) {};
              \edge[draw=none];
              \node[invis] (s5) {};
              \edge[draw=none];
              \node[invis] (s6) {};
              \edge[draw=none];
              \node[invis] (s7) {};
              \edge[draw=none];
              \node[invis] (s8) {};]
          \edge[draw=none];
             [.\node[invis] (s2) {};
              \edge[draw=none];
              \node[invis] (s9) {};
              \edge[draw=none];
              \node[invis] (s10) {};
              \edge[draw=none];
              \node[invis] (s11) {};
              \edge[draw=none];
              \node[invis] (s12) {};]
             \node[round] (s3) {};
             [.\node[round] (s4) {};
              \node[round] (s13) {};
              \node[round] (s14) {};]]
            
    \path pic[draw, -, angle radius=5mm] {angle = s3--s0--s4};
    \path pic[draw, -, angle radius=5mm] {angle = s13--s4--s14};
    
    \end{tikzpicture}}
    
    \caption{Progression of the cleanup algorithm on a partially explored policy tree. Black states and lines are in the currently returned policy, gray have been explored but are not returned. Crossed out states violate the risk bound.}
    \label{fig:Cleanup}
\end{figure}

Third, we note that the constraint in theorem~\ref{th:MajorTheorem} is not satisfied when sampling stops prematurely and there are outcomes of an action in the returned policy that are not sampled, as in figure~\ref{fig:Cleanup}. In that case, there may be a non-zero probability that an action in the policy leads to a state history that violates $\phi(f)$. To fix this, we add a new step to our algorithm that occurs after sampling is completed, which we call \emph{cleanup}. During cleanup, given in algorithm~\ref{alg:Cleanup}, all immediate outcomes of actions in the policy that have not yet been sampled are explored to ensure the lifetime reward up to those states satisfies $\phi(f)$. By doing so, theorem~\ref{th:MajorTheorem} applies to assert a risk bound over the actions in the policy that have been explored during tree search, as if the leaves of the search tree were terminal states. 

Cleanup is run by exploring all outcomes of the best policy found by UCT, and evaluating the lifetime reward up to unsampled states that follow immediately from actions in the policy. If $\phi(f)$ is found not to hold, the action leading to the state is deleted from the policy. The policy to take from a state history is set to the next best action from that state history, and the change in expected reward is propagated up the search tree. However, actions in the policy are \emph{not} changed during cleanup except when an action is deleted, in which case a different action from the same state history is chosen. This process is shown in figure~\ref{fig:Cleanup}, where an unexplored outcome of the current best policy is evaluated in (b), which results in an action being deleted. Instead of reevaluating the optimal action from the root state, the policy is only changed at the parent of the deleted action in (c), and unexplored outcomes of the new policy are evaluated. In this case, the alternative outcome also does not satisfy $\phi(f)$, so in (d), no valid actions remain from the parent of the deleted action, and the action that led to that state history is deleted. The immediate unexplored outcomes of the new policy are evaluated and found to be consistent with the risk bound. 

Even though deleting an action may change the optimal action further up the tree, we do not reevaluate alternative actions during cleanup for two reasons. First, we wish to avoid laborious computation during cleanup so that Vulcan terminates shortly after sampling is finished. While propagating changes in reward up a search tree is not computationally expensive, the change in the policy may require states to be evaluated all throughout the large search tree as the policy updates. Secondly, when sampling has not yet identified $\phi(f)$ not to hold, it is likely that many samples have been used to explore states that are close relatives to the violating state. Performing cleanup in the way we have detailed results in returned policies with more explored states deeper in the search tree.   

Regardless, it is not necessarily true that updating the policy further up the tree is a better approximation for the true policy up to the planning horizon, since we do not know whether high reward actions may occur after the violating state. In addition, the convergence of Vulcan to the policy $\pi^{*f}$ holds regardless of how cleanup is executed, since in the limit of a large number of samples, state histories violating $\phi(f)$ will be identified over the course of sampling.

\begin{algorithm}\label{alg:Vulcan}
\caption{Vulcan}
\KwIn{Start state $s_{0}$.}
\KwOut{Value function $\tilde{Q}(s_{0},\pi(s_{0}))$, policy $\pi$, which converges to $\pi^{*f}$.}
Initialize all $\tilde{Q}(h_{0:t},a)$, $N_{h_{0:t}}$, and $N_{h_{0:t},a}$ to 0\\
\While{\textup{run time} $\leq$ \textup{limit}}
      {\If{\textup{not Sample}$(s_{0})$}
          {\Return no solution}}
Cleanup$(s_{0})$ \\
\Return{$\tilde{Q}(s_{0},\pi), \pi$}
\end{algorithm}

\begin{algorithm}\label{alg:Sample}
\caption{Sample}
\KwIn{History to current state $h_{0:t}$.}
\KwOut{True if a sample to the planning horizon that respects the risk bound is found, false otherwise. Sets $\pi(h_{0:t})$.}
\If{$t = n$}
{\eIf{$ser(h_{0:n}) \leq \Delta(f(h_{0:n}))$}
     {Increment $N_{h_{0:t}}$ \\
      \Return true}
     {\Return false}}
\Loop{}
     {\If{\textup{no actions remain at} $h_{0:t}$}
         {\Return false}
      \eIf{$h_{0:t}$ \textup{has never been sampled}}
          {$a_{t} \leftarrow$ DefaultPolicy$(h_{0:t})$}
          {$a_{t} \leftarrow a_{uct}$}
      Select $s_{t+1}$ according to $p(s_{t+1}|h_{0:t},a_{t})$ \\
      \eIf{\textup{Sample}$(h_{0:t+1})$}
          {$N_{h_{0:t},a_{t}} \leftarrow \sum\limits_{h_{0:t+1}} N_{h_{0:t+1}}$ \\
           $N_{h_{0:t}} \leftarrow \sum\limits_{a_{t}} N_{h_{0:t},a_{t}}$ \\
           $\tilde{Q}(h_{0:t},a_{t}) \leftarrow \sum\limits_{h_{0:t+1}} \frac{N_{h_{0:t+1}}}{N_{h_{0:t},a_{t}}} \left( R(s_{t},a_{t},s_{t+1}) + \max\limits_{a'} \gamma \ \tilde{Q}(h_{0:t+1},a') \right)$ \\
           $\tilde{Q}(h_{0:t},\pi) \leftarrow \max\limits_{a} \tilde{Q}(h_{0:t},a)$ \\
           $\pi(h_{0:t}) \leftarrow \argmax\limits_{a} \tilde{Q}(h_{0:t},a)$ \\
           \Return true}
          {Delete $a_{t}$ at $h_{0:t}$ \\
           $N_{h_{0:t}} \leftarrow \sum\limits_{a_{t}} N_{h_{0:t},a_{t}}$}}
\end{algorithm}

\begin{algorithm}\label{alg:Cleanup}
\caption{Cleanup}
\KwIn{History to current state $h_{0:t}$.}
\KwOut{True if all outcomes of $\pi$ from current state history respect risk bounds, false otherwise. Sets policy $\pi(h_{0:t})$.}
\If{$h_{0:t}$ \textup{has never been sampled}}
   {\eIf{$ser(h_{0:t}) \leq \Delta(f(h_{0:t}))$}
        {\Return true}
        {\Return false}}
\If{$t = n$}
   {\Return true}
\Loop{}
     {\If{\textup{no actions remain at} $h_{0:t}$}
         {\Return false}
      \For{\textup{all safe outcomes of} $\pi(h_{0:t})$}
          {Cleanup$(h_{0:t+1})$}
      \eIf{\textup{all cleanup calls return true}}
          {$N_{h_{0:t},\pi(h_{0:t})} \leftarrow \sum\limits_{h_{0:t+1}} N_{h_{0:t+1}}$ \\
           $N_{h_{0:t}} \leftarrow \sum\limits_{a_{t}} N_{h_{0:t},a_{t}}$ \\
           $\tilde{Q}(h_{0:t},\pi) \leftarrow \sum\limits_{h_{0:t+1}} \frac{N_{h_{0:t+1}}}{N_{h_{0:t},\pi(h_{0:t})}} \left( R(s_{t},\pi(h_{0:t}),s_{t+1}) + \max\limits_{a'} \gamma \ \tilde{Q}(h_{0:t+1},a') \right)$ \\
           \Return true}
          {Delete $\pi(h_{0:t})$ from $h_{0:t}$ \\
           $\pi(h_{0:t}) \leftarrow \argmax\limits_{a} \tilde{Q}(h_{0:t},a)$}}
\end{algorithm}

\section{Experiments}

We examined the performance of Vulcan by testing it in two problem domains. The first may be viewed as a simplified multi-armed bandit problem with a limited number of actions \cite{gittins2011}, and we investigate this domain for small enough problem instances such that run time and optimality can be measured against existing methods. The second domain is illustrative of the exploration problems that Vulcan was built to solve, and describes a vehicle moving through a field described by a Gaussian Process \cite{williams2006}. Our second experiment shows the capability of Vulcan to deal with large state spaces without the need for strong heuristics.

All experiments were performed on a computer powered by an Intel 8-core i7 CPU with 4.00 GHz clock speed and 12 GB RAM allocated to the process.  

\subsection{Simplified Multi-Armed Bandit Domain}

In our multi-armed bandit domain, a player is faced with three machines, and must choose which machine to play at each action, subject to the following rules.

The reward for playing a machine is drawn from a two-point distribution with known outcomes, i.e. the machine returns reward $R_{1}$ with probability $p$, and reward $R_{2}$ with probability $1-p$, where $R_{1}$ and $R_{2}$ are known. The probability $p$ is not known, but is modeled in the Bayesian sense as also following a two-point distribution between two known values, meaning $p=p_{1}$ with probability $\theta$ and $p=p_{2}$ with probability $1-\theta$. The player begins with knowledge of $p_{1}$ and $p_{2}$ and an estimate for $\theta$.

After choosing a machine to play, the player immediately receives a randomly drawn reward from the distribution of the chosen machine. After receiving a reward, the player updates their belief state of $p$ according to the following Bayesian update rule. If the reward received after playing a given machine $i$ times is $R^{i}$ and the belief state after the $i$ measurements is characterized by $\theta^{i}$, then probability of a measurement is
\begin{equation*}
\begin{split}
    \mathbb{P}\left[ R^{i} \middle| \theta^{i-1} \right] &= \mathbb{P}\left[ R^{i} \middle| p_{1} \right] \mathbb{P}\left[p_{1} \middle| \theta^{i-1} \right] + \mathbb{P}\left[ R^{i} \middle| p_{2} \right] \mathbb{P}\left[p_{2} \middle| \theta^{i-1} \right] \\
    &= \mathbb{P}\left[ R^{i} \middle| p_{1} \right] \theta^{i-1} + \mathbb{P}\left[ R^{i} \middle| p_{2} \right] (1 - \theta^{i-1}),
\end{split}
\end{equation*}
while the update rule is 
\begin{equation*}
\begin{split}
    \theta^{i} &= \mathbb{P}\left[p_{1} \middle| R^{i}, \theta^{i-1} \right] \\
    &= \frac{\mathbb{P}\left[R^{i} \middle| p_{1} \right] \mathbb{P}\left[ p_{1} \middle| \theta^{i-1} \right]}{\mathbb{P}\left[ R^{i} \middle| \theta^{i-1} \right]} \\
    &= \frac{\mathbb{P}\left[R^{i} \middle| p_{1} \right] \theta^{i-1}}{\mathbb{P}\left[R^{i} \middle| p_{1} \right] \theta^{i-1} + \mathbb{P}\left[R^{i} \middle| p_{2} \right] (1 - \theta^{i-1})}.
\end{split}
\end{equation*}

Finally, a played machine may fail with a known risk of failure $r$. If any machine fails, zero reward is received for the action and the game immediately ends. If the player deems the risk of every machine to be too high, or instead all machines have been found to have low expected reward, the player may instead choose to end the game immediately as an action. Doing so has no risk, and grants a reward of 0.25 for every remaining action in the game ($0.25(n-t)$). 

The parameters for the specific machines we tested on are given in table~\ref{tab:bandit_parameters}. The player is forced to choose between a balanced machine with the highest and lowest rewards and intermediate risk (1), a relatively safe machine with low rewards but a belief that it is more likely to give the higher of its reward options (2), and a risky higher rewarding machine that is believed to be biased towards its low reward options (3). 

\begin{table}
    \centering
    \begin{tabular}{|c|c|c|c|c|c|c|}
    \hline
               & $R_{1}$ & $R_{2}$ & $p_{1}$ & $p_{2}$ & $\theta$ & $r$  \\
     \hline
     Machine 1 & 0 & 1 & 0.3 & 0.7 & 0.5 & 0.001 \\
     Machine 2 & 0.2 & 0.5 & 0.2 & 0.5 & 0.6 & 0.0005  \\
     Machine 3 & 0.4 & 0.6 & 0.3 & 0.6 & 0.3 & 0.0015 \\
     \hline
\end{tabular}
    \caption{Parameters for the three machines in the multi-armed bandit domain.}
    \label{tab:bandit_parameters}
\end{table}

This domain shows some of the aspects which motivated the development of Vulcan, including rewards that depend on the state history of the CCMDP, as well as the need to update a model after each measurement, which in this case is a relatively inexpensive Bayesian update. However, this case differs from a vehicle exploration problem in that the current state does not depend on the order of the measurements made, only the choices and outcomes. Vulcan generates a tree structure, and by modeling states in this domain as a tree we quantified the speedup of Vulcan for problems where the entire history is relevant (for example, with rewards that change in time). In addition, by modeling states as only dependent on the measurements and outcomes, creating a graph structure, we found that Vulcan still provides significant speedups over established methods, even when identical states are repeatedly generated.    


\subsubsection{VulcanFS Suboptimality}

Even without performing MCTS, VulcanFS finds solutions rapidly by enforcing the chance constraint through local conditions. The problem is then decoupled and Bellman updates can be performed at the expense of suboptimality in the solution. To characterize the suboptimality we solved the multi-armed bandit domain with various horizons (number of actions) using both VulcanFS and the MILP approach of~\cite{dolgov2005}, which generates the optimum deterministic policy. The VulcanFS rewards indicate the reward of the policy that Vulcan would eventually converge to. The problems used a risk bounding function of $\Delta(x) = 0.002 x$, with results shown in table~\ref{tab:bandit_reward}. The results suggest that the suboptimality is small, and remains small over a large number of problem horizons.

\begin{table}[htb]
    \centering
    \begin{tabular}{|c|c|c|c|}
    \hline
     Horizon & Optimal Reward & VulcanFS Reward & Suboptimality \\
     \hline
     2 & 0.9906 & 0.9906 & 0\% \\
     3 & 1.5280 & 1.4892 & 2.54\% \\
     4 & 2.0627 & 2.0167 & 2.23\% \\
     5 & 2.6068 & 2.5201 & 3.32\% \\
     6 & 3.1518 & 3.0686 & 2.64\% \\
     7 & 3.7006 & 3.5959 & 2.83\% \\
     8 & 4.2526 & 4.1334 & 2.80\% \\ 
     \hline
\end{tabular}
    \caption{Cumulative reward, and suboptimality for multi-armed bandit problems run for several horizons.}
    \label{tab:bandit_reward}
\end{table}

In order to verify that low suboptimality is not limited exclusively to the risk bounding function chosen in the example above, we ran VulcanFS against the MILP formulation for a continuum of risk bounding functions of the form $\Delta(x) = \alpha x$, for $\alpha$ in the range 0.0005 to 0.003, for a fixed horizon of 4 actions. Beyond these limits, there was no error in the VulcanFS solution. The form of the risk bounding function was limited to a proportional equation so that it could be encoded as a MILP. 

\begin{figure}[htbp]
    \centering
    \includegraphics[width=11cm]{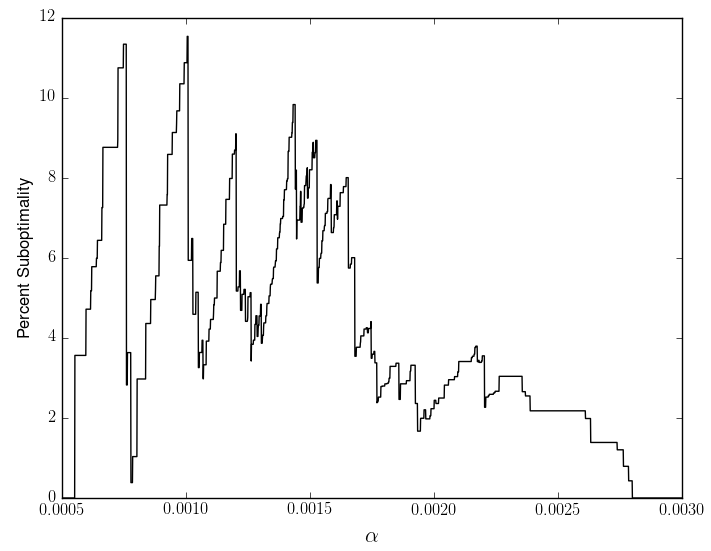}
    \vspace{-0.5cm}\caption{Suboptimality of VulcanFS for the multi-armed bandit problem with a horizon of 4 and a risk bounding function $\Delta(x) = \alpha x$.}
    \label{fig:bandit_suboptimality}
\end{figure}

The results are shown in figure~\ref{fig:bandit_suboptimality}, from which it appears that the suboptimality reduces as the risk bounding function becomes less restrictive. The regions in which the suboptimality exceeds 6\% were limited to relatively small regions of the test domain, and the average suboptimality over problems with nonzero suboptimality was 4.45\%. Run time was not found to be a strong function of the risk bounding function. 

\begin{figure}[htbp]
    \centering
    \includegraphics[width=11cm]{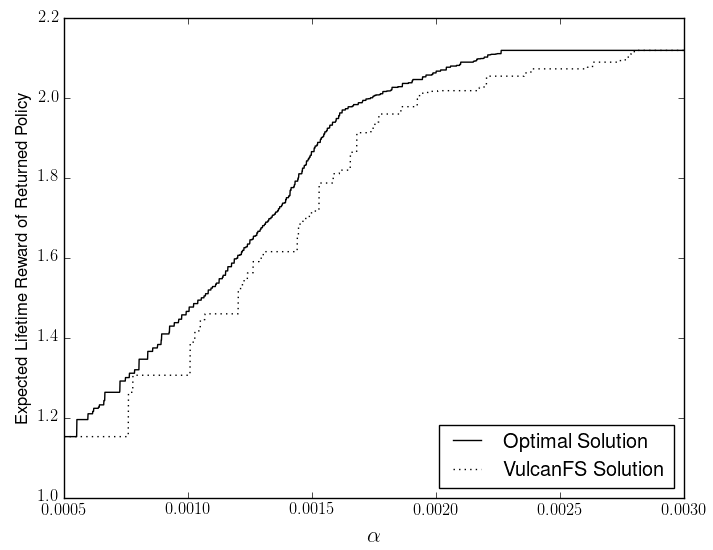}
    \vspace{-0.5cm}\caption{Expected rewards of the optimal policies and policies returned by VulcanFS for the multi-armed bandit problem with a horizon of 4 and a risk bounding function $\Delta(x) = \alpha x$.}
    \label{fig:bandit_solutions}
\end{figure}

The spikes in suboptimality result from increasing expected reward in the true optimal policy with $\alpha$, while the solution found by VulcanFS remains static. This is made clear by figure~\ref{fig:bandit_solutions}, which shows the expected reward of the true optimal policy and the VulcanFS solution. While small changes in $\alpha$ lead to increases in expected reward, VulcanFS continues to return the same solution over a larger range of functions until a better policy in $\Pi^{f}$ is found. As a result, the solution returned by VulcanFS shows more dramatic jumps, leading to a larger suboptimality immediately before increases in reward. As the risk bounding function becomes more lax, changes in the true optimal reward are less dramatic, and suboptimality decreases as a result. 

\subsubsection{VulcanFS Run Time for Increasing Problem Size}

\begin{figure}[htbp]
    \centering
    \includegraphics[width=11cm]{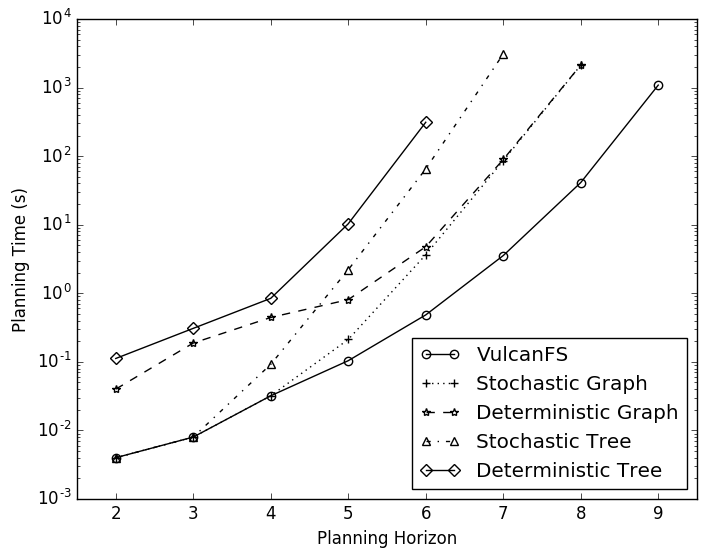}
    \vspace{-0.5cm}\caption{Planning time comparison of VulcanFS and LP based methods on the multi-armed bandit domain.}
    \label{fig:bandit_run_time}
\end{figure}

Examining the run time of VulcanFS is instructive because Vulcan should converge to the policy found by VulcanFS at least as quickly. In figure~\ref{fig:bandit_run_time} we show VulcanFS run time for increasing planning horizons in the multi-armed bandit domain. This was compared to the time taken to solve an equivalent CMDP allowing both stochastic policies (by LP) and deterministic policies (by MILP), and using a tree structure (repeated states) and a graph structure (without repeated states). In all tests, a risk bounding function of $\Delta(x) = 0.002 x$ was used. Since VulcanFS performs model updates and evaluates reward during run time, the times measured for the LP approaches include the time taken both to construct the LP and then to solve it.  

When attempting to find the deterministic policies for problems where transitions depend on the entire state history (deterministic tree), VulcanFS provides the most significant speedup. At a planning horizon of 6 actions, Vulcan provides a 600 fold decrease in run time, and enables much larger problems with larger planning horizons to be solved in realistic time frames. Even when stochastic policies are allowed and there are many ways to reach a state (stochastic graph), VulcanFS still reaches a solution over 50 times faster than an LP based method, despite forming a tree structure and without any optimization for repeating states.    


\subsubsection{Vulcan Convergence}

\begin{figure}[htb]
    \centering
    \includegraphics[width=12cm]{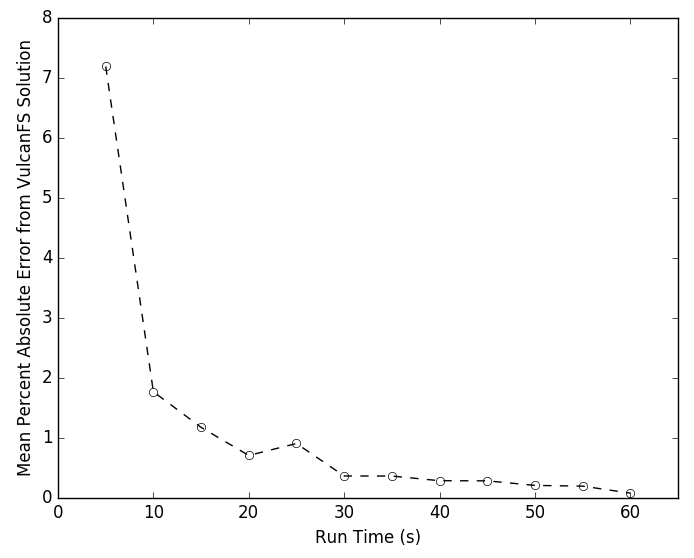}
    \vspace{-0.5cm}\caption{Mean absolute error in Vulcan solution as a function of run time. VulcanFS requires approximately 1080 seconds to find a solution.}
    \label{fig:bandit_convergence}
\end{figure}

Since Vulcan uses MCTS to run faster than VulcanFS, we investigated its convergence to the policy found by VulcanFS for the the bandit problem with a horizon of 9 actions. No LP based method was able to solve this problem in under two hours, and VulcanFS required approximately 18 minutes. For every 5 seconds between 5 and 60 seconds, Vulcan was run 60 times. The average absolute error from the VulcanFS solution is shown in figure~\ref{fig:bandit_convergence}. With 60 seconds of run time (less than 6\% of the run time of VulcanFS) Vulcan was able to achieve a mean error of 0.08\%. At that run time, the exact same policy as VulcanFS was found 90\% of the time. 

\subsubsection{Vulcan Performance Against Heuristic Search}

The previous sections have shown Vulcan's speedup over methods that require an explicit representation of the entire state space. In heuristic search methods, heuristics guide the search, leading to an optimal policy without exploring the entire state space. The strength of Vulcan is in the fact that it tends to find policies quickly even if no strong heuristic is known. However, we now show that Vulcan produces policies an order of magnitude faster than the heuristic search method RAO* \cite{santana2016} for large problems in our multi-armed bandit domain, even when it is possible to develop a strong heuristic. 

RAO* requires a reward heuristic $\psi_{Q}(s_{t},a_{t})$ which overestimates the expected cumulative reward that can be gained from a state by taking a given action, and an execution risk heuristic $\psi_{er}(s_{t},a_{t})$ which underestimates the execution risk from a state given an action. In the multi-armed bandit domain, a simple upper bound is the expected reward of the best possible machine multiplied by the number of actions. The best possible machine is machine 1 operating with $p = 0.3$, leading to an expected reward of 0.7,
\begin{equation*}
    \psi_{Q}(s_{t},a_{t}) = 0.7 (n - t).
\end{equation*}

An underestimate of the execution risk is the immediate risk of the action to be taken, 
\begin{equation*}
    \psi_{er}(s_{t},a_{t}) = r(s_{t},a_{t}).
\end{equation*}

For relatively small problems (up to a planning horizon of 7) RAO* ran approximately twice as fast as the time taken for Vulcan to converge to its final solution, which equaled up to one second time difference. At these scales, it appeared that the practical computational overhead of progressing through the search tree by random sampling at every iteration and tracking value functions at each node was significant compared to the speedup gained by random sampling. At a planning horizon of 8 run times were comparable, and at a planning horizon of 9 Vulcan ran approximately twice as fast as RAO*.

\begin{figure}[htb]
    \centering
    \includegraphics[width=12cm]{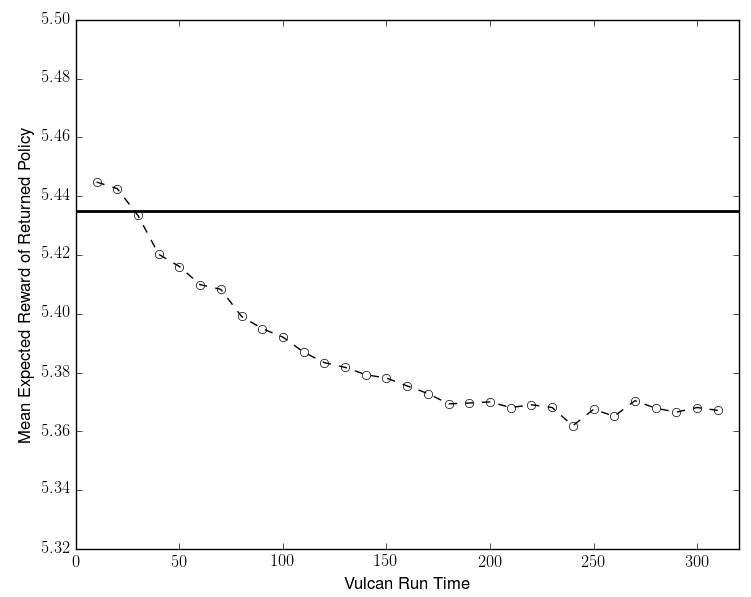}
    \vspace{-0.5cm}\caption{Mean expected reward of policies output by Vulcan as a function of run time for the multi-armed bandit domain with 10 actions. The expected reward of the optimum policy is shown by the bold line, which was found by RAO* in 3515.9 seconds.}
    \label{fig:bandit_rao_convergence}
\end{figure}

At a planning horizon of 10, the problem was large enough to show a noticeable difference in run time. The experiment was run with a constant risk bound of $\Delta = 0.0125$, and RAO* took 3515.9 seconds to find the optimal solution. Vulcan was run 100 times for run times of every 10 seconds between 10 and 310 seconds. The average rewards of the policies are shown in figure~\ref{fig:bandit_rao_convergence}. 

Unlike unconstrained MCTS, figure~\ref{fig:bandit_rao_convergence} shows that the average expected reward does not always increase with time with Vulcan. For short run times, the state space is not fully explored, so the returned policies satisfy the risk bound only up to the states explored. This can lead to overestimates of the reward, since unlikely states that violate $\phi(f)$ have not been found. Further sampling tends to decrease the expected reward as unfeasible policies are recognized and eliminated.

\begin{figure}[htb]
    \centering
    \includegraphics[width=12cm]{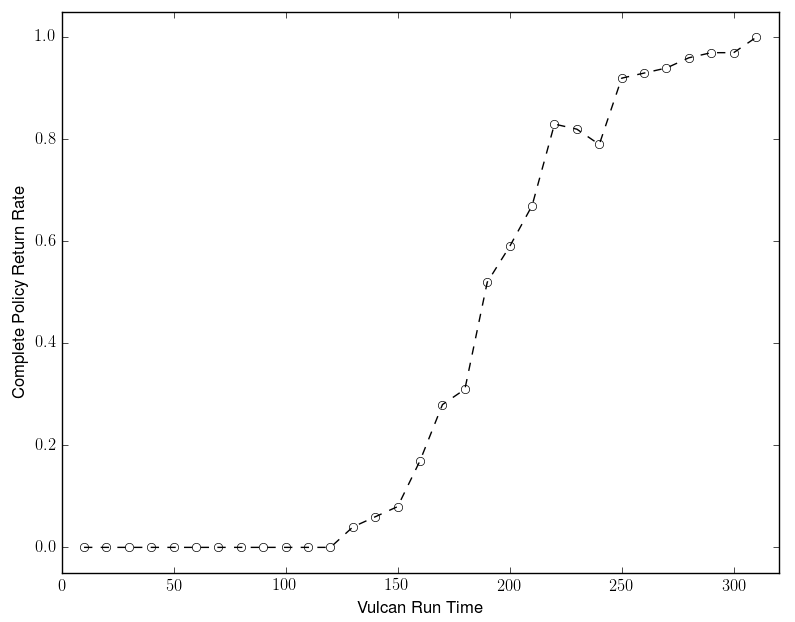}
    \vspace{-0.5cm}\caption{Complete policy return rate as a function of run time for the multi-armed bandit domain with 10 actions. A complete policy assigns an action to all reachable state histories up to the planning horizon.}
    \label{fig:bandit_rao_completion}
\end{figure}

The policies found with the short run times are typically incomplete, in the sense that they do not include actions for all reachable outcomes up the planning horizon and only satisfy the risk bound up to the states that have been explored. Figure~\ref{fig:bandit_rao_completion} shows the complete policy return rate as a function of run time. With 310 seconds of run time or greater, a complete policy was always returned. These complete policies satisfy the risk bound, are found 11.3 times faster than RAO*, and have a mean suboptimality of only 1.2\%. 

\subsection{Exploration of a Gaussian Process}

Our second test domain concerns a vehicle that is tasked with maximizing samples taken through an environment described by a Gaussian Process (GP) (for GP details, see for example \cite{williams2006}). 

In our domain, the vehicle's state is described by a gaussian distribution $\mathcal{N}(\mu,\Sigma)$. The mean state $\mu$ is constrained to a rectangular grid, and each action corresponds to the choice of one of the 8 adjacent locations to which the mean location will move. Meanwhile, after each action, the covariance at time step $t$ grows according to
\begin{equation*}
    \Sigma_{t} = \Sigma_{t-1} + \Sigma_{w} 
\end{equation*}
for known $\Sigma_{w}$, as in \cite{ono2013}. In our experiment, we used
\begin{equation*}
    \Sigma_{0} = \begin{bmatrix}
                 0.005 & 0 \\
                 0 & 0.005
                 \end{bmatrix}
\end{equation*}
\begin{equation*}
    \Sigma_{w} = \begin{bmatrix}
                 0.0001 & 0 \\
                 0 & 0.0001
                 \end{bmatrix}.
\end{equation*}

Known obstacles exist in the environment, and the risk considered in the domain is the probability of collision with those obstacles. Using the methods of \cite{ono2013}, the probability of collision with a single obstacle is estimated as the probability that the gaussian distribution passes one of the boundaries of the obstacle. At a given time step, the total probability of failure is conservatively estimated as the sum of the probabilities of collision with all obstacles. 

After moving to a new location, a sample with no error is taken from a GP with known hyperparameters, but no prior measurements is taken before the mission begins. For simplicity, each sample is modeled as taken at the vehicle's mean location. If the location has not been previously visited, a reward is received equal to the value of the sample. After a sample is taken, the Gaussian Process model from the previous state is updated to include the new measurement. The vehicle is tasked with moving through the environment to maximize the sum of its samples.

The Gaussian Process model is described by a mean function $m(\mathbf{x})$ and squared exponential covariance kernel $k(\mathbf{x}_{1},\mathbf{x}_{2})$ of the form
\begin{equation*}
    m(\mathbf{x}) = 1 + \begin{bmatrix} 0.05 & 0.05 \end{bmatrix} \mathbf{x}
\end{equation*}
\begin{equation*}
    k(\mathbf{x}_{1},\mathbf{x}_{2}) = 0.16 \exp \left( - \frac{1}{8} \lVert \mathbf{x}_{1} - \mathbf{x}_{2} \rVert^{2} \right).
\end{equation*}
These hyperparameters encourage the vehicle to move to the top right of the environment in order to maximize its samples, but it must bypass obstacles in the way in order to do so. The GP describes a gaussian distribution over measurements, so the discrete outcomes that were considered in the CCMDP were chosen by Gauss-Hermite quadrature with degree 4 \cite{hildebrand1987}. 

The exploration domain shows all of the features that motivated the design of Vulcan. The possible outcomes of future actions are a function of all observations that are taken during the mission, and since the risk of entering a state is also a function of time, the order those measurements are taken in affects the probability of failure. As a result, the risk and reward accumulated depends explicitly on the state history. Prediction of the outcomes of a measurement requires an inversion of a Gaussian Process at each state, which makes reward evaluation relatively expensive. 

The problem we ran on this domain began with a fixed initial location and measurement, and allowed 9 additional actions to be taken. With 8 available actions and 4 possible outcomes of each action, the problem has approximately $3.5 \times 10^{13}$ unique state histories.

\begin{figure}
    \centering
    \includegraphics[width=12.5cm]{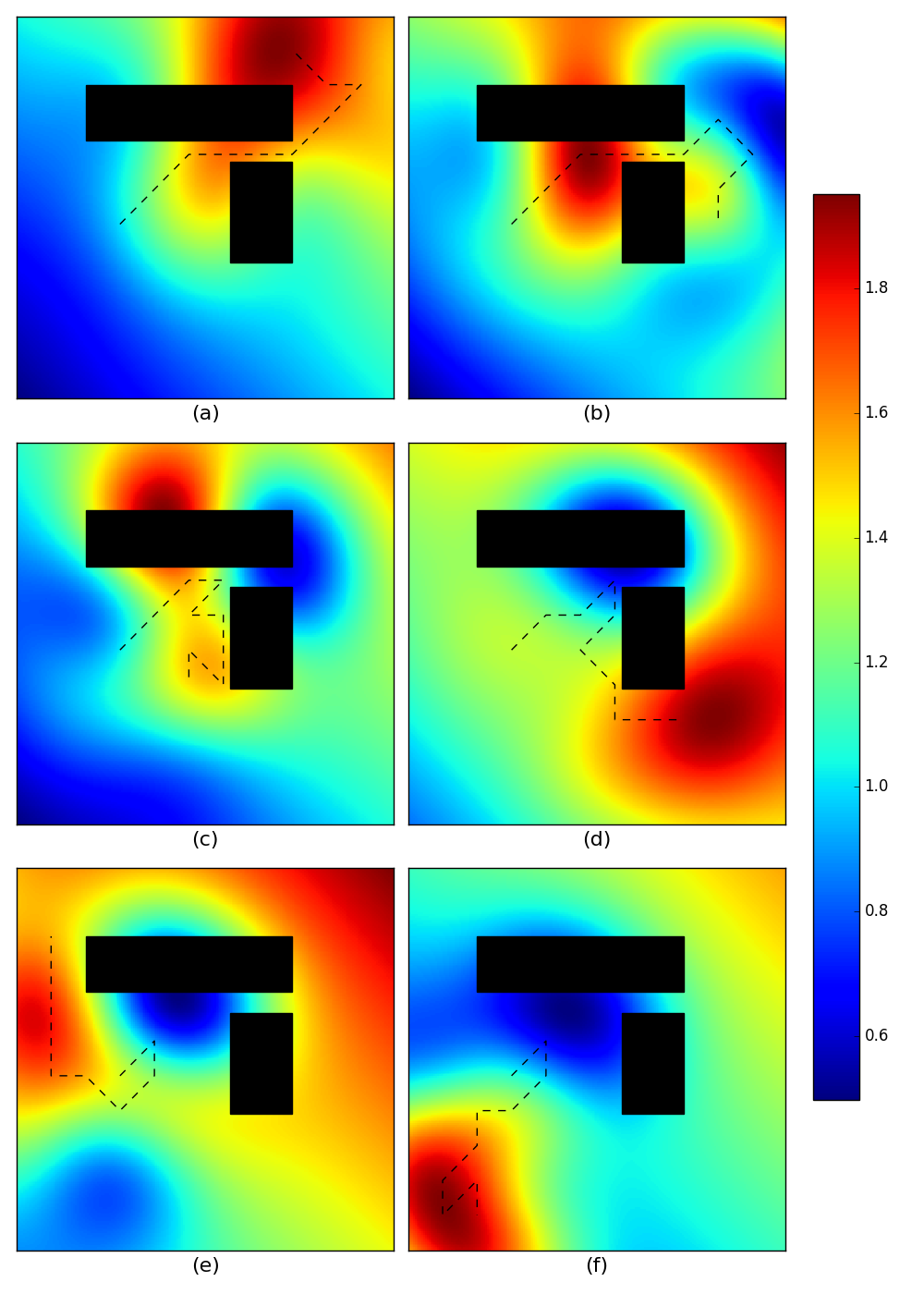}
    \caption{Six possible outcomes of a single policy generated for the Gaussian Process exploration problem with a risk bounding function $\Delta(x) = (1 - \exp(-0.4x)) (0.015 + 0.001 x)$.}
    \label{fig:GP_Path_1}
\end{figure}

Vulcan was run on the Gaussian Process exploration domain with 9 actions and allowed 180 seconds of run time. A risk bounding function of the form $\Delta(x) = (1 - \exp(-0.4x)) (0.015 + 0.001 x)$ was used, which begins at the origin and approaches a straight line for large x. This function was chosen to show Vulcan's capability to handle nonlinear concave risk bounding functions, and the specific parameters in the function, GP, and covariance evolution were chosen so that a high risk path between the two obstacles was permissible only for sufficiently high reward, which did not always occur. A small selection of the potential outcomes of the policy found are shown in figure~\ref{fig:GP_Path_1}, but they demonstrate the interplay between risk and reward that underlies the reasoning performed by Vulcan. 

The trajectories shown in figure~\ref{fig:GP_Path_1} (a) and (b) travel through a narrow gap between obstacles. In doing so, there is an increased probability of failure, but this is justified by high reward predictions from the Gaussian Process model. In the case of (a), the explored field is high near the gap, and it is reasoned that it will continue to be high nearby. In the case of (b) the additional risk is justified by large observations before the gap is reached, so traveling through the gap is permitted even though high reward is not found. 

In (c) and (d) the vehicle moves towards the gap, encouraged by the increasing mean in that direction, but low measurements are found immediately before the gap. In the case of (c) higher reward is expected to be found downwards, and relatively high measurements justify staying close to the obstacle. In (d) the reward is low at the gap and it is no longer worth the risk to travel through the gap, even though the environment's mean increases in that direction.  

Finally, in (e) and (f), an immediate low measurement suggests that the risk of continuing to explore near the obstacles outweighs the expected reward, and the vehicle moves away from the obstacles. In both cases, the vehicle follows the trends observed in the data to move towards a region of high reward.


\section{Conclusions}

In this paper we presented Vulcan, which uses Monte Carlo Tree Search to produce approximately optimal policies for chance constrained MDPs subject to a concave nondecreasing risk bounding function. Whereas previous methods for CCMDPs have been limited to problems with millions of states when strong heuristics are unavailable because the entire policy is coupled through the chance constraint, in Vulcan we decompose the chance constraint into constraints that are placed on individual state histories. By doing so, standard approaches for unconstrained MDPs can be efficiently applied, with a small degree of suboptimality introduced into the final solution. The application of MCTS allows policies to be found for problems with state spaces that are too large and computationally expensive to generate in full. This is particularly important when considering CCMDPs where rewards depend on the entire state history up to a given state.

Using Vulcan, we solved approximately optimal policies for CCMDPs that are orders of magnitude larger than those handled in the literature without heuristics. Experimentally, we showed that Vulcan finds solutions tens to hundreds of times faster than methods based on linear programming, and that convergence happens rapidly without the need to fully explore the state space. Additionally, Vulcan was used to find complete policies 11.3 times faster than heuristic forward search for a sufficiently large problem, and in all experiments the returned policies were found to have a mean expected reward that differed from the true optimal policy by a few percent. We then applied Vulcan to an exploration problem with approximately $10^{13}$ unique states, and found a policy that appropriately balanced risk against reward in 180 seconds.

\section{Acknowledgements}

The authors would like to acknowledge and thank the Exxon Mobil Corporation for their financial support (grant EM09079).

\vskip 0.2in
\bibliography{main}
\bibliographystyle{theapa}

\end{document}